\documentclass[final]{cvpr}

\usepackage{times}
\usepackage{epsfig}
\usepackage{graphicx}
\usepackage{amsmath}
\usepackage{amssymb}

\usepackage{color}
\usepackage{cite}
\usepackage{tabularx}
\usepackage{adjustbox}
\usepackage{threeparttable}
\usepackage{multirow}
\usepackage{subfigure}
\usepackage{caption}
\usepackage{booktabs}
\usepackage[normalem]{ulem}
\usepackage{outlines}
\usepackage[dvipsnames]{xcolor}
\usepackage{microtype}

\DeclareMathOperator*{\argmin}{argmin}

\newcolumntype{s}{>{\raggedleft\arraybackslash}X}

\newcommand{\bx}{\mathbf{x}}

\newcommand{\mLM}{\mathcal{L}_M}

\newcommand{\bw}{\mathbf{w}}

\newcommand{\Supp}{Appendix}
\newcommand{\supp}{appendix}

\usepackage{enumitem}
\setlist[itemize]{leftmargin=*, topsep=3pt}

\usepackage{ifthen}
\newboolean{arxiv}

\usepackage[pagebackref=true,breaklinks=true,colorlinks,bookmarks=false]{hyperref}

\newcommand{\ourmodel}{\textsc{MP3}}
\newcommand{\ourdataset}{\textsc{UrbanExpert}}

\def\cvprPaperID{7124} %
\def\confYear{CVPR 2021}

\newcommand{\cutequationup}{\vspace*{-2pt}}
\newcommand{\cutequationdown}{\vspace*{-2pt}}
\newcommand{\cutsectionup}{\vspace*{-4pt}}
\newcommand{\cutsectiondown}{\vspace*{-2pt}}
\newcommand{\cutsubsectionup}{\vspace*{-2pt}}
\newcommand{\cutsubsectiondown}{\vspace*{-2pt}}
\newcommand{\cutsubsubsectionup}{\vspace*{-2pt}}
\newcommand{\cutsubsubsectiondown}{\vspace*{-1pt}}
\newcommand{\cutparagraphup}{\vspace*{-2pt}}
\newcommand{\cutparagraphdown}{\vspace*{-1pt}}
\newcommand{\cuthalfcaptionup}{\vspace*{-6pt}}
\newcommand{\cutcaptionup}{\vspace*{-20pt}}
\newcommand{\cutcaptiondown}{\vspace*{-10pt}}
\begin{document}

\setboolean{arxiv}{true}

\title{MP3: A Unified Model to Map, Perceive, Predict and Plan}

\author{
   Sergio Casas\thanks{Denotes equal contribution}$~^{, 1, 2}$, Abbas Sadat $^{*, 1}$, Raquel Urtasun$^{1, 2}$\\
   $\text{Uber ATG}^1$, $\text{University of Toronto}^2$ \\
   $\texttt{\{sergio, urtasun\}@cs.toronto.edu, abbas.sadat@gmail.com}$
}

\maketitle

\begin{abstract}
   High-definition maps (HD maps) are a key component of most modern self-driving systems due to their valuable semantic and geometric information.
Unfortunately, building HD maps has proven hard to scale due to their cost as well as the requirements  they impose in the localization system that has to work everywhere with centimeter-level accuracy. %
Being able to drive without  an HD map would be very beneficial to scale self-driving solutions as well as to increase the failure tolerance of existing ones (e.g., if localization fails or the map is not up-to-date). 
Towards this goal, we propose \ourmodel{}, an end-to-end approach to mapless\footnote{
    We note that by \textit{mapless} we mean without HD maps. A coarse road network like the ones available in off-the-shelf services such as Google Maps or OpenStreetMap is assumed available for routing towards the goal.
} driving 
where the input is raw  sensor data and a high-level command (e.g., turn left at the intersection).
\ourmodel{} predicts intermediate representations in the form of an online map and the current and future state of dynamic agents, 
and exploits them in a novel neural motion planner to make interpretable decisions taking into account uncertainty. 
We show that our approach is significantly safer, more comfortable, and can follow commands better than the baselines in challenging long-term closed-loop simulations, as well as when compared to an expert driver in a large-scale real-world dataset.

\end{abstract}

\cutsectionup
\section{Introduction}
\cutsectiondown

Most modern self-driving stacks require up-to-date high-definition (HD) maps that contain  rich semantic information necessary for driving such as the topology and location  of the lanes, crosswalks, traffic lights,  intersections as well as the traffic rules for each lane (e.g., unprotected left, right turn on red, maximum speed). 
These maps are a great source of knowledge that simplify  the  perception and  motion forecasting tasks, as the online inference process  has to mainly focus   on   dynamic objects (e.g., vehicles, pedestrians, cyclists). Furthermore, the use of HD maps significantly increases the safety of motion planning as knowing the lane topology and geometry eases the generation of potential trajectories for the ego-vehicle that adhere to the traffic rules. 
In addition, progressing towards a specific goal is much simpler when the desired route is defined as a sequence of lanes to traverse.

\begin{figure}[t]
    \centering
    \includegraphics[width=\columnwidth]{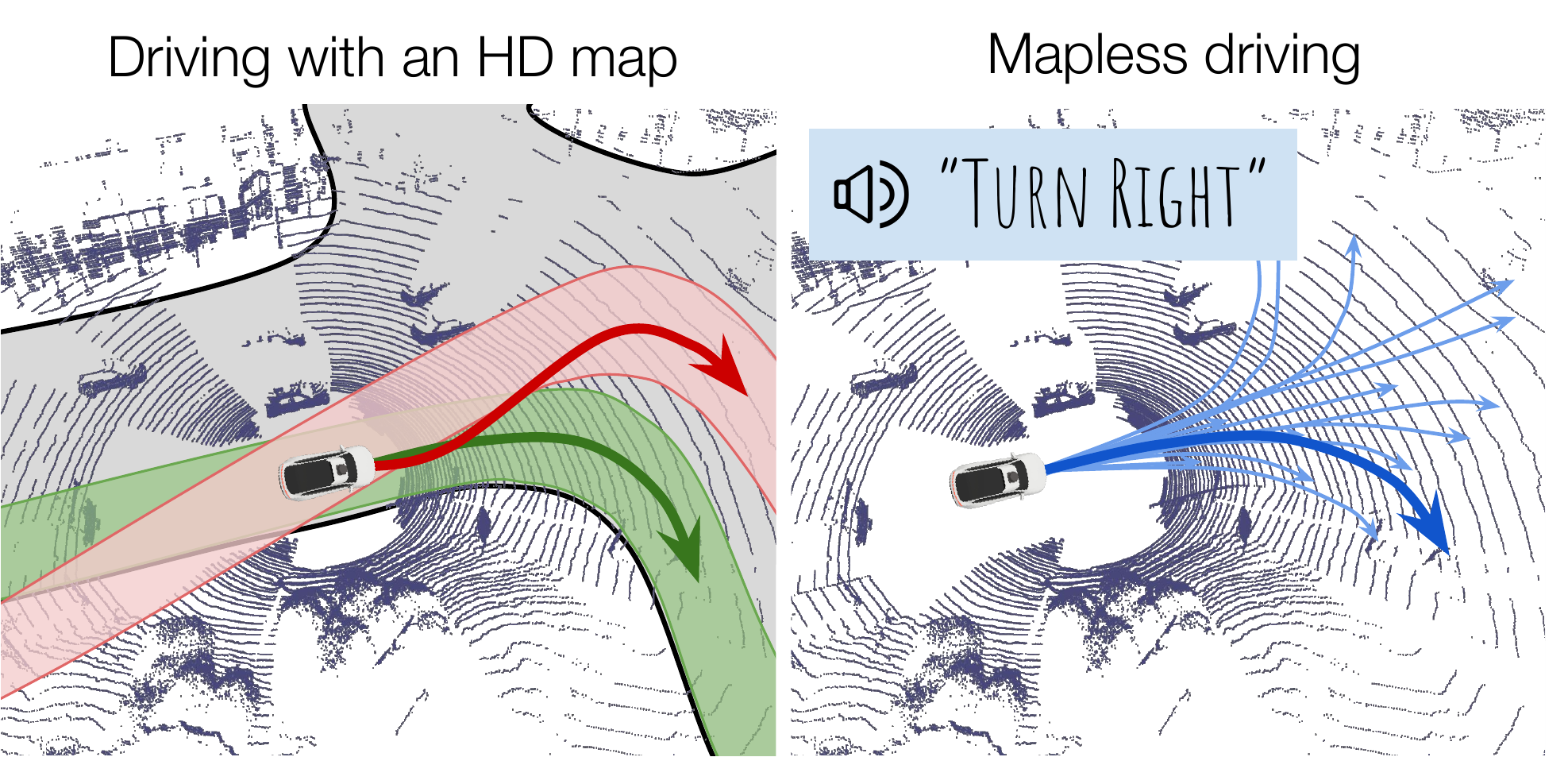}
    \cutcaptionup
    \caption{
        Left: a localization error makes the SDV follow a wrong route when using an HD map, driving into traffic. Right: mapless driving can interpret the scene from sensors and achieve a safe plan that follows a high-level command. 
    }
    \cutcaptiondown
    \label{fig:teaser}
\end{figure}

Unfortunately, building HD maps has proven  hard to scale due to the complexity and cost of generating the maps and maintaining them. 
Furthermore, the heavy reliance on HD maps introduces very demanding requirements for the localization system, which needs to work at all times with centimeter-level accuracy or else unsafe situations like Fig.~\ref{fig:teaser} (left) might arise. 
This motivates the development of mapless technology, which can serve as the fail-safe in the case of localization failures or outdated maps, and potentially unlock self-driving at scale at a much lower cost.

Self-driving without HD maps is a very challenging task. Perception can no longer rely on the prior that is more likely to find vehicles on the road and pedestrians on the sidewalk. Motion forecasting of dynamic objects   becomes even more challenging without having access to the lanes that vehicles typically follow or the location of crosswalks for pedestrians.
Most importantly, the search space to plan a safe maneuver for the SDV goes from narrow envelopes around the lane center lines \cite{werling2010optimal, ajanovic2018search, sadat2019jointly, sadat2020perceive} to the full set of dynamically feasible trajectories as depicted in Fig.~\ref{fig:teaser} (right). Moreover, without a well-defined route as a series of lanes to follow, the goal that the SDV is trying to reach needs to be abstracted into high-level behaviors such as going straight at an intersection, turning left or turning right \cite{codevilla2018end}, which require taking different actions depending of the context of the scene.

Most mapless approaches \cite{pomerleau1989alvinn, bojarski2016end, codevilla2018end, hawke2019urban, philion2020lift}, focus on imitating the controls of an expert driver (e.g., steering and acceleration), without providing intermediate interpretable representations that can help explain the self-driving vehicle decisions. 
Interpretability is of key  importance in a safety-critical system particularly if a bad event was to happen. Moreover, the absence of a mechanism to inject structure and prior knowledge makes these methods  very brittle to distributional shift \cite{ross2011reduction}.  
While methods that perform online mapping to obtain lane  boundaries or lane center lines  have been proposed 
\cite{bai2018deep, homayounfar2018hierarchical, garnett20193d, guo2020gen,philion2020lift}, they either are overly simplistic (e.g., assume lanes are close to parallel to the direction of travel), have only been demonstrated in highway scenarios which are much simpler than city driving, have not been shown to work when coupled with any existing planner, or involve information loss through discrete decisions such as confidence thresholding to output the candidate lanes. 
The latter is safety-critical as a lane can be completely missed in the worst case, and  it makes it difficult to incorporate uncertainty about the static environment in motion planning, which is importance to reduce risk. %

To address these challenges, we propose an  end-to-end approach to mapless driving that is interpretable, does not incur any information loss, and reasons about uncertainty in the intermediate representations.
In particular, we propose a set of probabilistic spatial layers to model the static and dynamic parts of the environment. 
The static environment is subsumed in a planning-centric online map which captures information about which areas are drivable and which ones are reachable given traffic rules. 
The dynamic actors are captured in a novel occupancy flow that provides occupancy and velocity estimates over time.
The  motion planning module then leverages these representations without any postprocessing. 
It utilizes observational data to retrieve dynamically feasible trajectories, predicts a spatial mask over the map to estimate the route given an abstract goal, and leverages the online map and occupancy flow directly as cost functions for explainable, safe plans.
We showcase that our approach is significantly safer, more comfortable, and can follow commands better than a wide variety of baselines in challenging closed-loop simulations, as well as when compared to an expert driver in a large-scale real-world dataset.

\begin{figure*}[t]
   \vspace{-15pt}
   \centering
   \includegraphics[width=\textwidth]{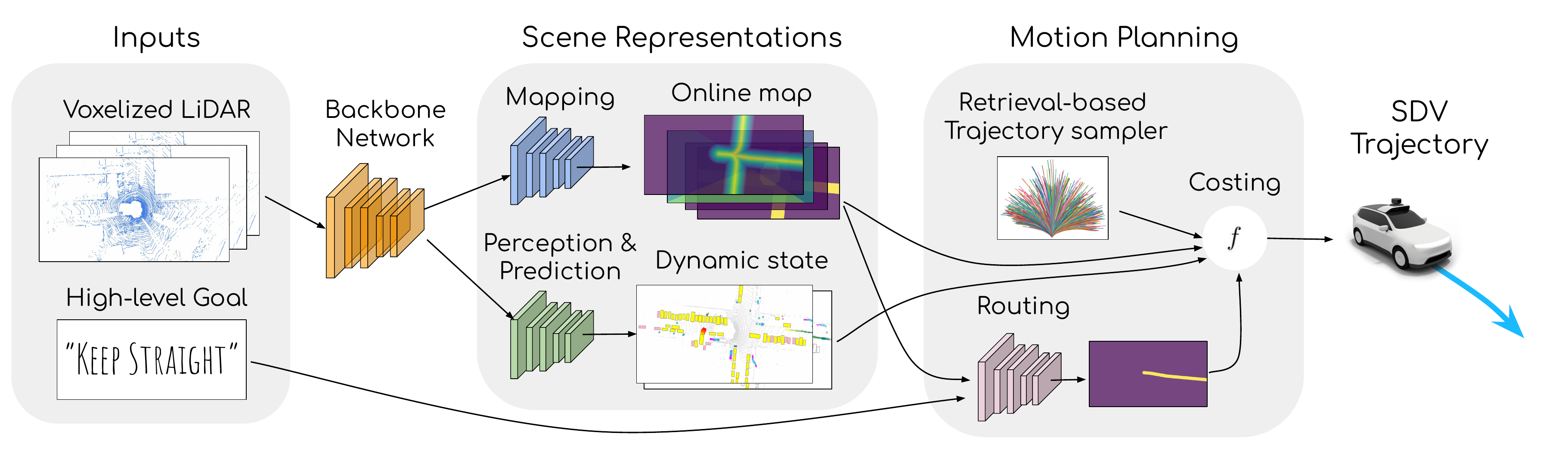}
   \cutcaptionup
   \caption{MP3 predicts probabilistic scene representations that are leveraged in motion planning as interpretable cost functions} 
   \cutcaptiondown
   \label{fig:overview}
\end{figure*}

\cutsectionup
\section{Related Work}
\cutsectiondown
We cover previous works on online mapping, perception and prediction, and motion planning, particularly analyzing their fitness to the downstream task of end-to-end driving.

\cutparagraphup
\paragraph{Online Mapping:}
\cutparagraphdown
While there are many offline mapping approaches \cite{mattyus2017deeproadmapper, bastani2018roadtracer, homayounfar2019dagmapper}, these rely on satellite imagery or multiple passes through the same scene with a data collection vehicle to gather dense information, and often involve a human-in-the-loop. For these reasons, such approaches are not suitable for mapless driving.
As a consequence, predicting map elements online has recently been proposed. In \cite{garnett20193d, guo2020gen} a network is presented to directly predict the 3D layout of lanes in a traffic scene from a single image.
Conversely to the methods above, \cite{bai2018deep} argues that accurate image estimates do not translate to precise 3D lane boundaries, which are the input required by modern motion planning algorithms. To tackle this, LiDAR and camera are used to predict estimates of ground height and lanes directly in 3D space.
Alternatively, \cite{homayounfar2018hierarchical} proposes a hierarchical recurrent neural network for extraction of structured lane boundaries from LIDAR sweeps. 
Notably, all the works above are geared toward highway traffic scenes and involve discrete decisions that could be unsafe when driving as they lose valuable uncertainty information.
Contrary to these methods, we leverage dense representations of the map that do not involve information loss and are suitable for use in the motion planner as interpretable cost functions.

\cutparagraphup
\paragraph{Perception and Prediction:}
\cutparagraphdown
Most previous works perform object detection \cite{yang2018pixor, lang2019pointpillars,zhou2019end, meyer2019lasernet, frossard2020strobe} and actor-based prediction to reason about the current and future state of a driving scene. As there are multiple possible futures, these methods either generate a fixed set of trajectories \cite{luo2018faf, casas2018intentnet, casas2019spatially, phan2019covernet, chai2019multipath, zhao2019multi, Liang_2020_CVPR, casas2020importance, li2020end}, draw samples to characterize the distribution \cite{rhinehart2018r2p2, tang2019multiple, casas2020implicit} or predict temporal occupancy maps \cite{jain2019discrete, ridel2020scene, liang2020garden}. 
However, these pipelines can be unsafe since the detection stage involves confidence thresholding and non-maximum suppression which can remove unconfident detections of real objects.
In robotics, occupancy grids at the scene-level (in contrast to actor-level) have been a popular representation of free space. Different from the methods above, \cite{elfes1989using, thrun2003learning} estimate occupancy probability of each grid-cell independently using range sensor data. 
More recently, \cite{hoermann2018dynamic} directly predicts an occupancy grid to replace object detection, but it does not predict how the scene might evolve in the future.
\cite{sadat2020perceive} improves over such representation by adding semantics as well as future predictions. However, there is no way to extract velocity from the scene occupancy, which is important for motion planning.
While \cite{wu2020motionnet} considers a dense motion field, their parameterization cannot capture multi-modal behaviors.
We follow the philosophy of \cite{elfes1989using, thrun2003learning, hoermann2018dynamic, sadat2020perceive, wu2020motionnet} in predicting scene-level representations, but propose an improved occupancy flow parameterization that can model multi-modal behavior and provides a consistent temporal motion field.

\cutparagraphup
\paragraph{Motion Planning:}
\cutparagraphdown
There is a vast literature on end-to-end approaches for self-driving.
The pioneering work of \cite{pomerleau1989alvinn} proposes to use a single neural network that directly outputs a driving control command.
Subsequent to the success of deep learning, direct control based methods have advanced with deeper networks, richer sensors, and scalable learning methods~\cite{bojarski2016end,kendall2018learning,codevilla2018end,muller2018driving}. 
Although simple and general, such methods of directly generating control command from sensor data may have  stability and robustness issues \cite{codevilla2019exploring}. 
More recently, cost map-based approaches have been shown to adapt better to challenging environments, which recover a trajectory by looking for local minima on the cost map. The cost map may be parameterized as a simple linear combination of hand crafted costs \cite{sadat2019jointly,fan2018auto}, or in a general non-parametric form \cite{zeng2019neural}.
To bridge the gap between interpretability and expressivity, \cite{sadat2020perceive} proposed 
a model that leverages supervision to learn an interpretable nonparametric occupancy that can be directly used in motion planner, with hand-crafted sub-costs.
In contrast to all methods above which rely on an HD map, \cite{ma2019navigatewithoutlocalize} proposes to output a navigation cost map without localization under a weakly supervised learning environment. This work, however, does not explicitly predict the static and dynamic objects and hence lacks safety and interpretability.
Similarly, \cite{banzhaf2019nonholonomic} 
improves sampling in complex driving environments without the consideration of dynamic objects, and is only demonstrated in simplistic static scenarios.
In contrast, our autonomy model leverages retrieval from expert demonstrations to achieve an efficient trajectory sampler that does not rely on the map, predicts a spatial route based on the probabilistic map predictions and a high-level driving command, and stays safe by exploiting an interpretable dynamic occupancy field as a summary of the scene free space and motion.

\cutsectionup
\section{Interpretable Mapless Driving}
\cutsectiondown
In this section, we introduce our  end-to-end approach to self-driving that operates directly on raw sensor data. Importantly, our model produces intermediate representations that are designed for safe planning,  decision-making and interpretability. 
Our interpretable representations estimate the current and future state of the world around the SDV, including  the unknown map as well as the current and future location and velocity of dynamic objects.
In the remainder of this section, we first describe  our  backbone network that extracts meaningful geometric and semantic features from the raw sensor data. We then introduce our intermediate interpretable representations, and show how they can be exploited to plan  maneuvers that are safe, comfortable, and explainable. An overview of our model can be seen in Fig.~\ref{fig:overview}

\cutsubsectionup
\subsection{Extracting Geometric and Semantic Features}
\cutsubsectiondown
Our model exploits a history of LiDAR point clouds to extract rich geometric and semantic features from the scene over time.
Following~\cite{luo2018faf}, we voxelize $T_p$=10 past LiDAR point clouds in bird's eye view (BEV), equivalent to 1 second of history, with a spatial resolution of $a=0.2$ meters/voxel.
We exploit odometry to compensate for the SDV motion,  thus voxelizing all point clouds in a common coordinate frame.
Our region of interest is $W$=140m long (70m front and behind of the SDV), $H$=80m wide (40 to each side of the SDV), and $Z$=5m tall. 
Following \cite{casas2018intentnet}, we concatenate height and time along the channel dimension to avoid using 3D convolutions or a recurrent model, thus saving memory and computation. 
The result is a 3D tensor of size $(\frac{H}{a}, \frac{W}{a}, \frac{Z}{a} \cdot T_p)$, which is the input to our backbone network. This network combines ideas from \cite{yang2018pixor,casas2018intentnet} to extract geometric, semantic and motion information about the scene.
More details can be found in the \supp{}.

\cutsubsectionup
\subsection{Interpretable Scene Representations}
\begin{figure}[t]
    \centering
    \vspace{-10pt}
    \begin{tabular} {c@{\hspace{.5em}}c}
        \raisebox{-0.5\height}{\includegraphics[width=0.45\columnwidth, trim={4.5cm, 4.0cm, 4.5cm, 4.0cm}, clip]{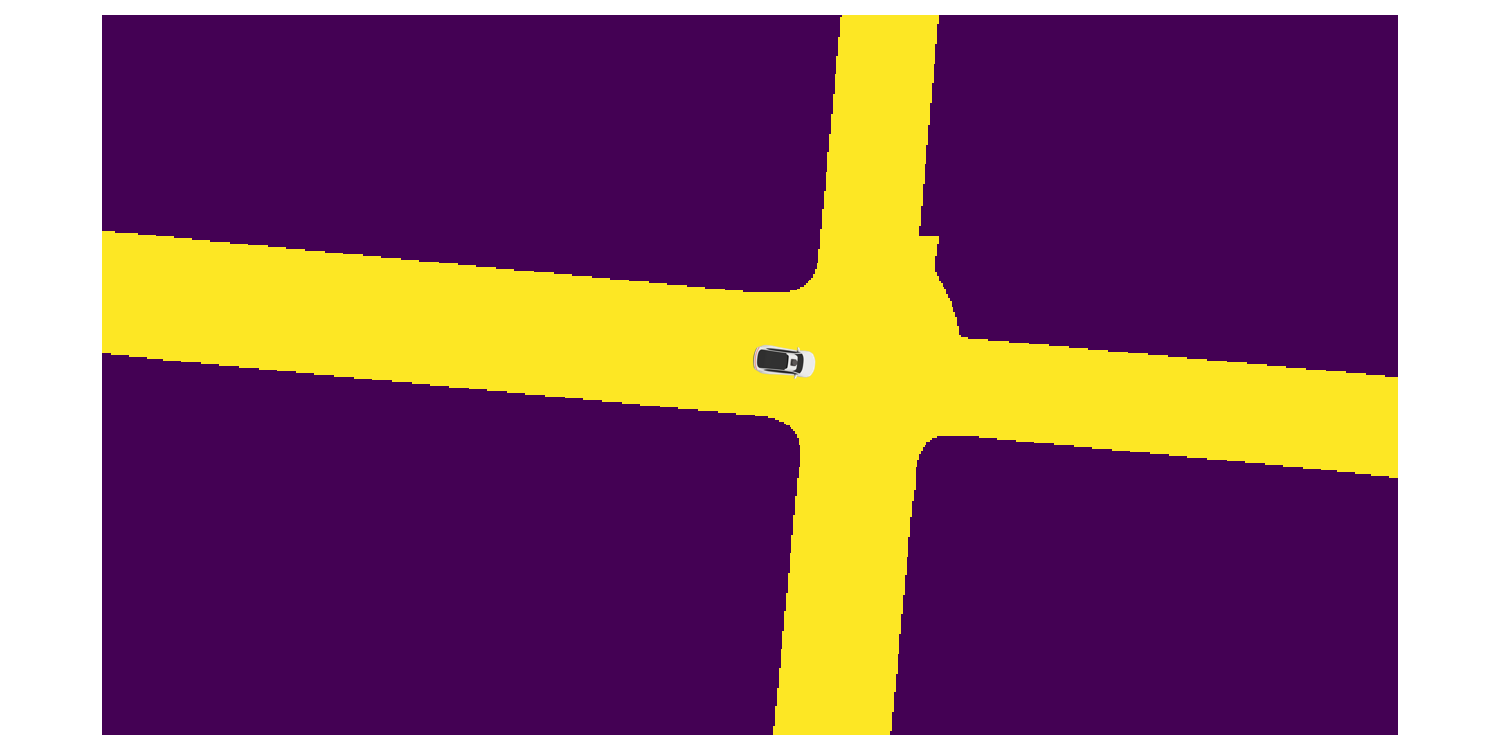}} &
        \raisebox{-0.5\height}{\includegraphics[width=0.45\columnwidth, trim={4.5cm, 4.0cm, 4.5cm, 4.0cm}, clip]{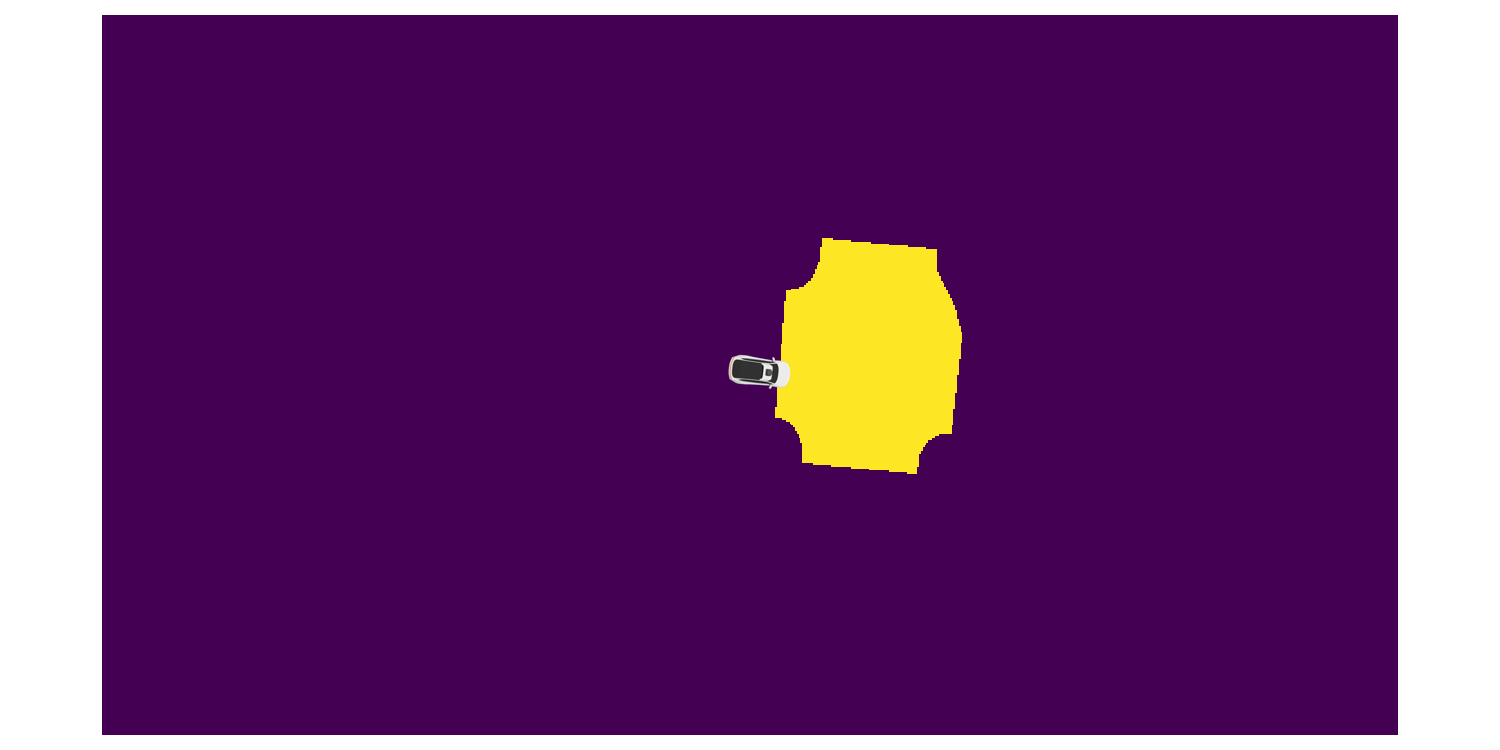}} \\
        Drivable area & Intersections \\
        \raisebox{-0.5\height}{\includegraphics[width=0.45\columnwidth, trim={4.5cm, 4.0cm, 4.5cm, 4.0cm}, clip]{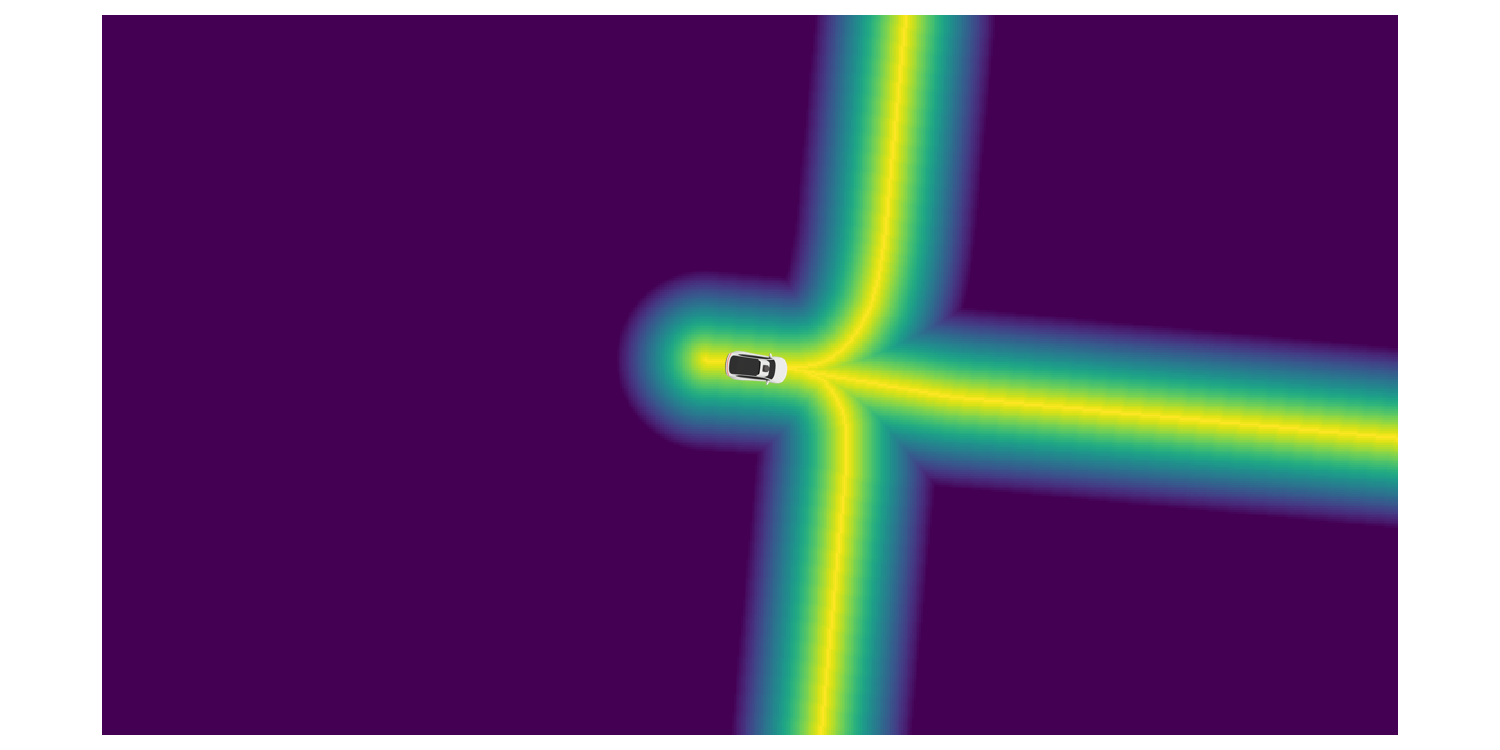}} \vspace{.5em} &
        \raisebox{-0.5\height}{\includegraphics[width=0.45\columnwidth, trim={4.5cm, 4.0cm, 4.5cm, 4.0cm}, clip]{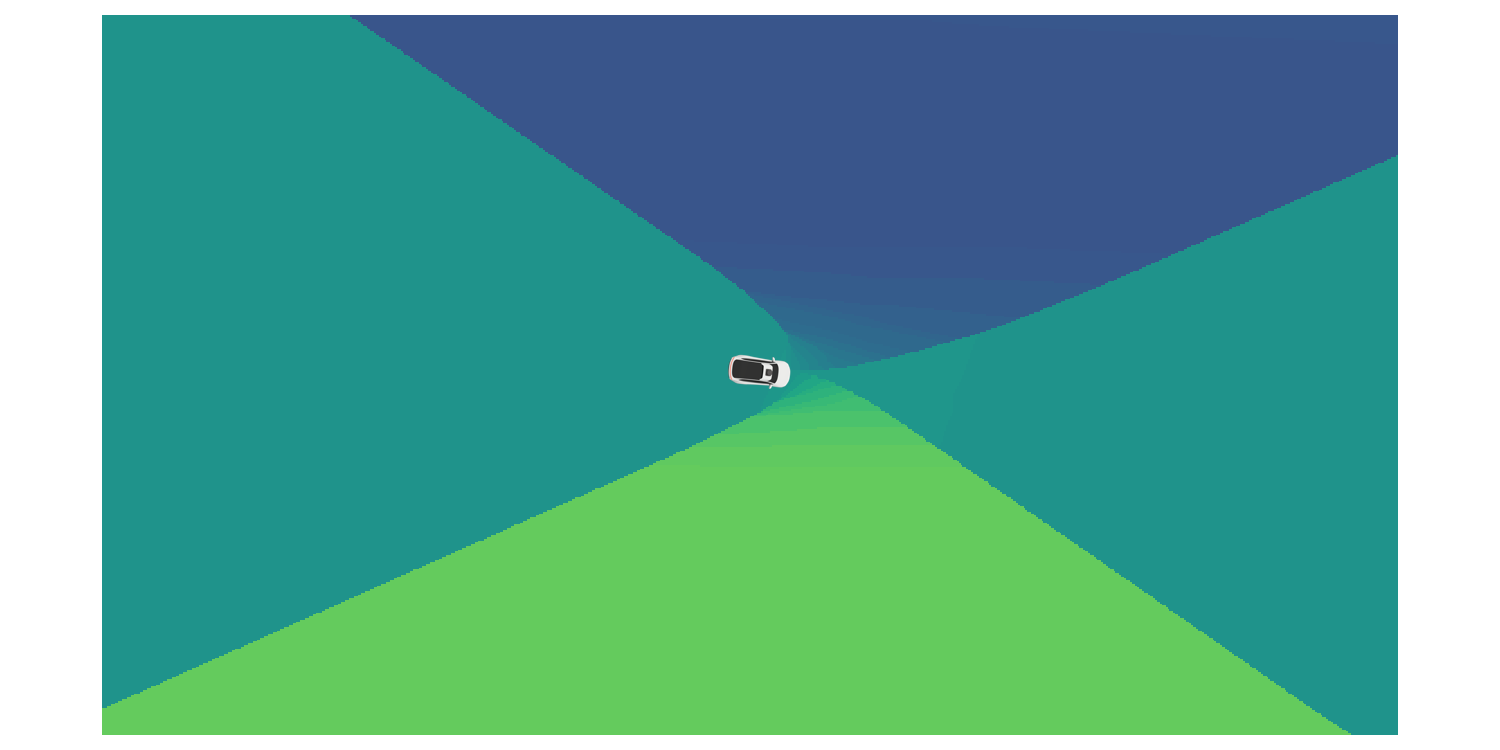}} \\
        Reachable Distance Transform & Reachable Angle \\
        \raisebox{-0.5\height}{\includegraphics[width=0.45\columnwidth, trim={4.5cm, 4.0cm, 4.5cm, 4.0cm}, clip]{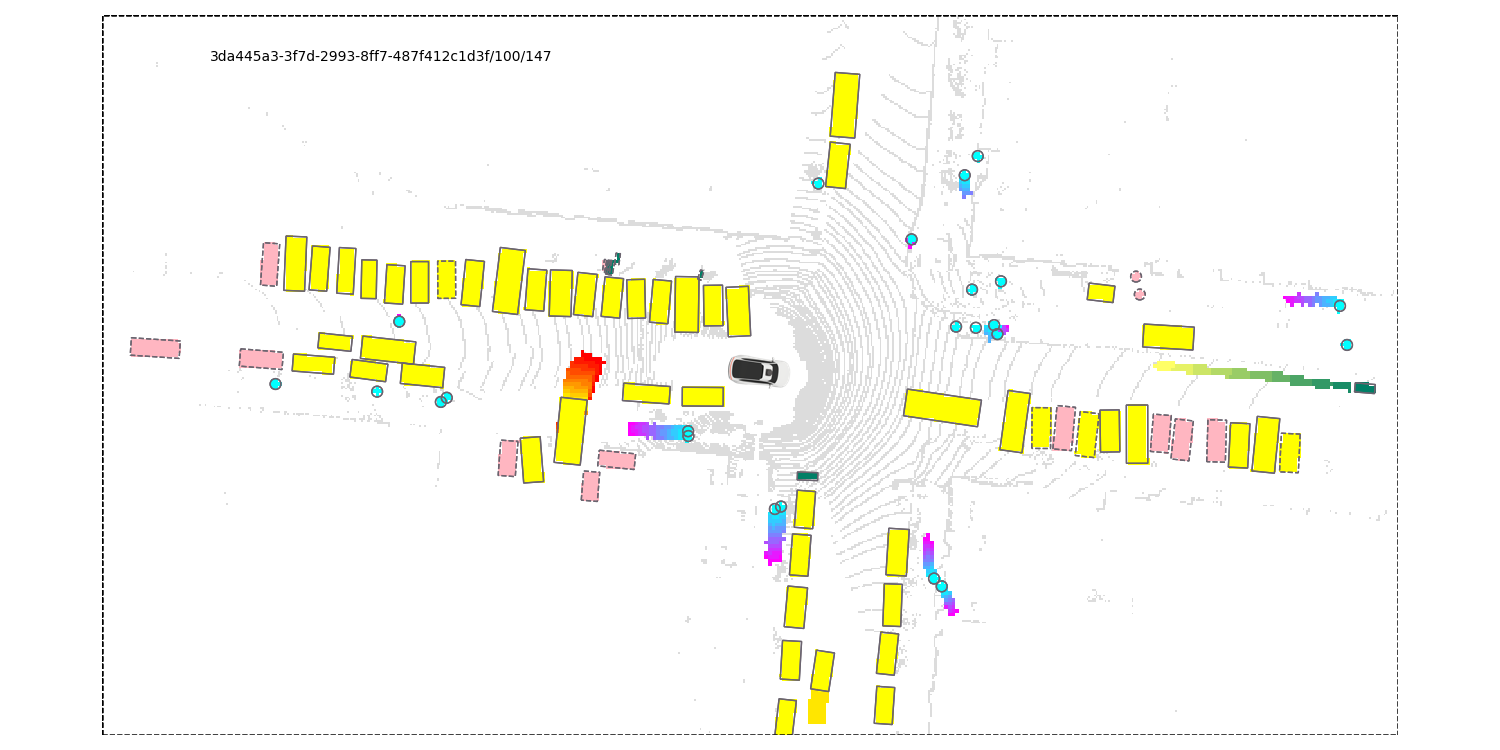}} &
        \raisebox{-0.5\height}{\includegraphics[width=0.45\columnwidth, trim={4.5cm, 4.0cm, 4.5cm, 4.0cm}, clip]{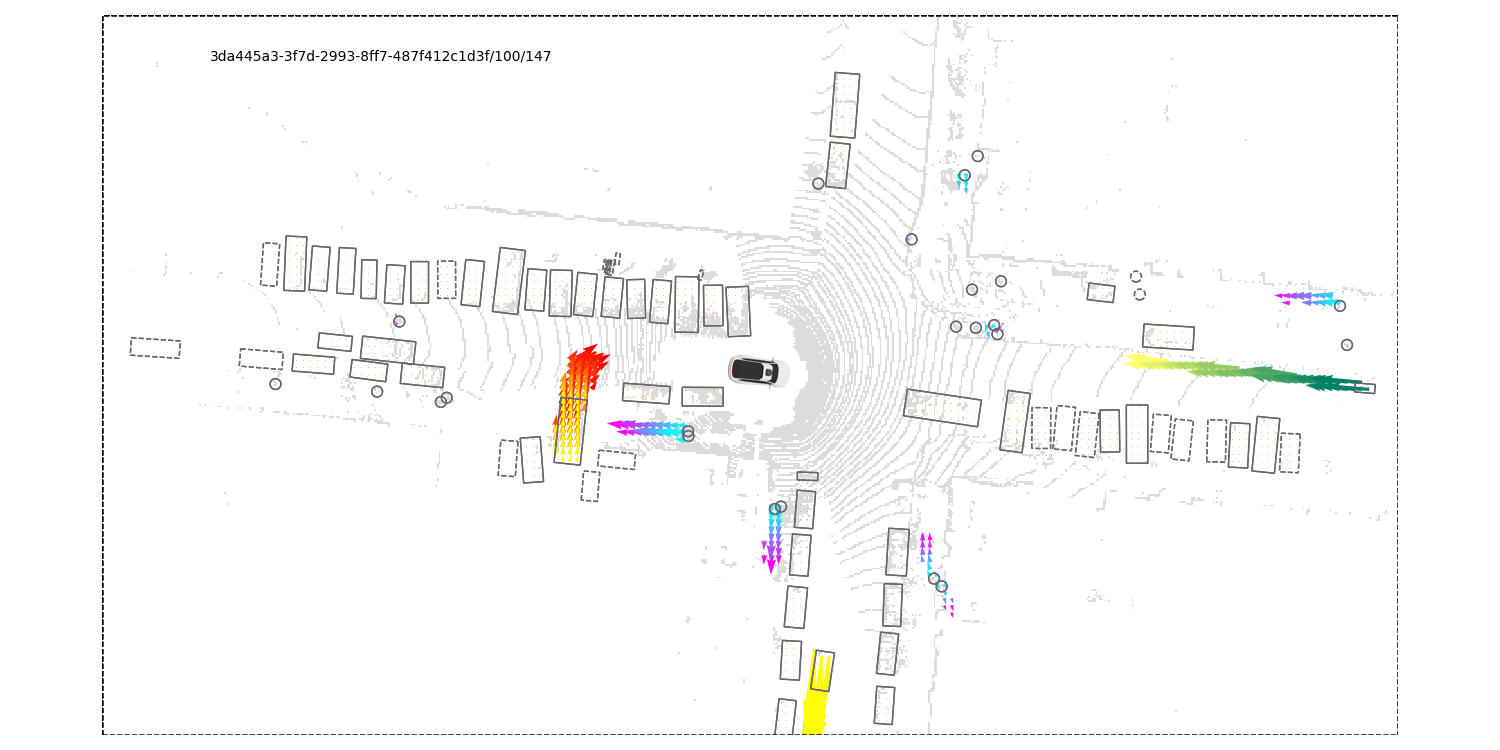}}
        \vspace{.5em} \\
        Occupancy & Temporal Motion Field
    \end{tabular}
    \cuthalfcaptionup
    \caption{\textbf{Interpretable Scene representations}. For occupancy and motion, we visualize all time steps and classes in the same image to save space, differentiating with colors. %
    }
    \cutcaptiondown
    \label{fig:semantic_layers}
\end{figure}

\cutsubsectiondown
Human drivers are able to successfully navigate complex road topologies with high-density of traffic by exploiting their prior knowledge about traffic rules and social behavior such as the fact that vehicles should drive on the road, close to a lane centerline,  in the direction of traffic and should not collide with other actors.
Since we would like to incorporate such prior knowledge into the decisions of the SDV, and these to be explainable through interpretable concepts, it is important to predict intelligible representations of the static environment, which  we refer here as an {\it online map}, as well as the dynamic objects position and velocity into the future, captured in our {\it dynamic occupancy field}. We refer the reader to  Fig.\ref{fig:semantic_layers} for an example of these representations.
Since the predicted online map and dynamic occupancy field are not going to be perfect due to limitations in the sensors,  occlusions and the model, it is important to reason about  uncertainty to assess the risk of each possible decision the SDV might take. Next, we first describe the semantics in our interpretable representation of the world, and then introduce our probabilistic model.

\cutparagraphup
\paragraph{Online map representation:}
\cutparagraphdown
In order to drive safely it is useful to reason the following elements in BEV:
\begin{itemize}[itemsep=1pt]
    \item \textit{Drivable area}: 
    Road surface (or pavement) where vehicles are allowed to drive, bounded by the curb.
    \item \textit{Reachable lanes}:
    Lane center lines (or motion paths) are defined as the canonical paths vehicles travel on, typically in the middle of 2 lane markers.
    We define the reachable lanes  as the subset of motion paths the SDV can get to without breaking any traffic rules.
    When planning a trajectory, we would like the SDV to stay close to these reachable lanes and drive aligned to their direction.
    Thus, for each pixel in the ground plane we predict the unsigned distance to the closest reachable lane centerline, truncated at 10 meters, as well as the angle of the closest reachable lane centerline segment. 
    \item \textit{Intersection}: 
    Drivable area portion where traffic is controlled via traffic lights or traffic signs.
    Reasoning about this is important to handle stop/yield signs and traffic lights. For instance, if a traffic light is red, we should wait to enter the intersection. 
    Following \cite{rhinehart2018deep}, we assume a separate camera-based perception system detects the traffic lights and recognizes their state as this is not our focus.
\end{itemize}

\begin{figure}[t]
    \vspace{-20pt}
    \centering
    \includegraphics[width=\columnwidth]{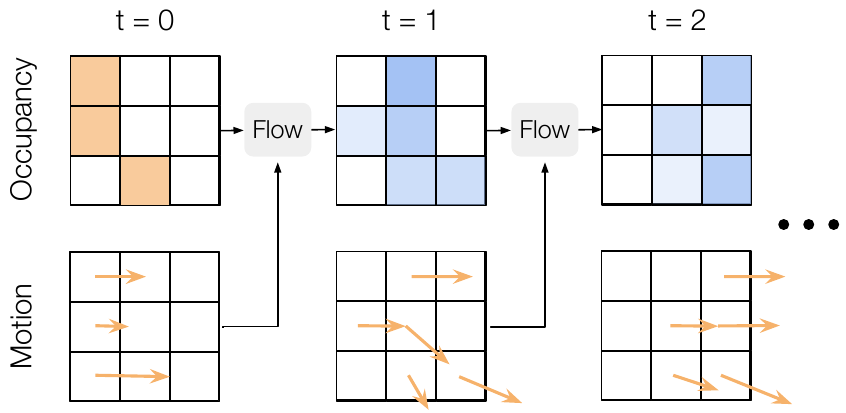}
    \cutcaptionup
    \caption{
        The motion field warps the occupancy over time. Transparency denotes probability.
        Color differences the {\color{Melon} predicted layers by the network} and the {\color{Cerulean} future occupancy}. 
        We depict the particular case of unimodal motion $(K=1)$.
    }
    \cutcaptiondown
    \label{fig:cartoon_occupancy_flow}
\end{figure}
\cutparagraphup
\paragraph{Dynamic occupancy field:}
\cutparagraphdown
Another critical aspect to achieve safe self-driving is to understand which space is occupied by dynamic objects and how do these move over time.
Many accurate LiDAR-based object detectors have been proposed \cite{yang2018pixor, lang2019pointpillars,zhou2019end, meyer2019lasernet} to localize dynamic obstacles followed by a motion forecasting stage \cite{luo2018faf, rhinehart2018r2p2, chai2019multipath, tang2019multiple, casas2020implicit} to predict the future state of each object. However, all these methods contain unsafe discrete decisions such as confidence thresholding and non-maximum suppression (NMS) that can eliminate low-confidence predictions of true objects resulting in unsafe situations.
\cite{sadat2020perceive}  proposed a probabilistic way to measure the likelihood of a collision for a given SDV maneuver by exploiting a non-parametric spatial representation of the world. This computation is agnostic to the number of objects. 
However, this representation does not provide velocity estimates, and thus it is not amenable to car-following behaviors and speed-dependent safety buffer reasoning.
Moreover, the decision making algorithm cannot properly reason about interactions, since for a given future occupancy its origin cannot be traced back.

In contrast, in this paper we propose an occupancy flow parameterized by the occupancy of the dynamic objects at the current state of the world and a temporal motion field into the future that describes how objects move (and in turn their future occupancies), both discretized into a spatial grid on BEV with a resolution of 0.4 m/pixel, as depicted in Fig.~\ref{fig:cartoon_occupancy_flow}: 
\begin{itemize}[itemsep=1pt]
    \item \textit{Initial Occupancy}: a BEV grid cell is active (occupied) if its center falls in the interior of a polygon given by an object shape and its current pose.
    \item \textit{Temporal Motion Field}: defined for the occupied pixels at a particular time into the future.  Each occupied pixel motion is represented with a 2D BEV velocity vector (in m/s).
    We discretize this motion field into $T=11$ time steps into the future (up to 5s, every 0.5s).

\end{itemize}

Since the SDV behavior should be adaptive to objects from different categories (e.g., extra caution is desired around vulnerable road users such as pedestrians and bicyclists), we consider vehicles, pedestrians and bikes as separate  classes, each with their own occupancy flow.
\cutparagraphup
\paragraph{Probabilistic Model:}
\cutparagraphdown
We would like to reason about uncertainty in our online map and dynamic occupancy field. 
Towards this goal, we model each semantic channel of the online map $\mathcal{M}$ as a collection of independent variables per BEV grid cell. This assumption makes the model very simple and efficient. %
To simplify the notation, we use the letter $i$ to indicate a spatial index on the grid instead of two indices (row, column) from now on.
We model each BEV grid cell in the drivable area and intersections channels as Bernoulli random variables, $\mathcal{M}^A_i$ and $\mathcal{M}^I_i$ respectively, as we consider a grid cell is either part these elements or not. We model the truncated distance transform to the reachable lanes centerline $\mathcal{M}^D_i$ as a Laplacian, which we empirically found to yield more accurate results than a Gaussian, and the direction of the closest lane centerline in the reachable lanes $\mathcal{M}^{\theta}_i$ as a Von Mises distribution since it has support between $[\pi,  \pi]$. 

We model  the occupancy of dynamic objects $\mathcal{O}^c$  for  each class $c \in \{\text{vehicle, pedestrian, bicyclist}\}$  as a collection of Bernoulli random variables $\mathcal{O}^c_{t, i}$, one  for each spatio-temporal index $t, i$. 
Since an agent future behavior is highly uncertain and multi-modal (e.g., a vehicle going straight vs. turning right), we model the motion for each class at each spatio-temporal location as a categorical distribution $\mathcal{K}^c_{t, i}$ over $K$ BEV motion vectors $\{\mathcal{V}^c_{t, i, k} : k \in 1 \dots K\}$. Here, each motion vector is parameterized by the continuous velocity in the x and y directions in BEV.
To compute the probability of future occupancy  under our probabilistic model, we first define the probability of occupancy flowing from location $i_1$ to location $i_2$ between two consecutive time steps $t$ and $t+1$ as follows:
\cutequationup
{\small
\begin{align*}
p(\mathcal{F}^c_{(t, i_1) \to (t+1, i_2)}) = \sum_k p(\mathcal{O}^c_{t, i_1}) p(\mathcal{K}^c_{t,i_1}=k) p(\mathcal{V}^c_{t, i_1, k}=i_2)
\end{align*}
}
where $p(\mathcal{V}_{t, i_1, k}=i_2)$ distributes the mass locally and is determined via bilinear interpolation  if $i_2$ is among the 4 nearest grid cells to the head of the continuous motion vector, and 0 for all other cells, as depicted in Fig.~\ref{fig:cartoon_occupancy_flow}. 
With this definition, we can easily calculate the future occupancy iteratively, starting from the occupancy predictions at $t=0$. 
This parameterization  ensures consistency by definition between future motion and future occupancy, and provides an efficient way to query how does some particular initial occupancy evolve over time, which will be used for interaction and right-of-way reasoning in our motion planner.
Specifically, to get the occupancy that flows into cell $i$ at time $t+1$ from all cells $j$ at time $t$, we can simply compute the probability that no occupancy flow event occurs, and take its complement
\cutequationup
{\small
\begin{align*}
p(\mathcal{O}^c_{t+1, i}) = 1 - \prod_{j}\big( 1 - p(\mathcal{F}^c_{(t, j) \to (t+1, i)})\big)
\end{align*}
}
\cutequationdown
We point the reader to the \supp{} for further details on the mapping and perception and prediction network architecture.

\cutsubsectionup
\subsection{Motion Planning} \label{sec:planning}  
\cutsubsectiondown

The goal of the motion planner is to generate trajectories that are safe, comfortable and progressing towards the goal. 
We design a sample-based motion-planner in which a set of kinematically-feasible trajectories are generated and then evaluated using a learned scoring function. 
The scoring function utilizes the probabilistic dynamic occupancy field to encode  the safety of the possible maneuvers encouraging cautious behaviors that avoid occupied regions, and maintain a safe headway to the occupied area in front of the SDV.
The probabilistic layers in our online map are used in the scoring function  to ensure the SDV is  driving on the drivable area, close to the lane center and in the right direction, being cautious in uncertain regions, and driving towards the goal specified by the input high-level command.
The planner  evaluates all the sampled trajectories  in parallel and selects the trajectory with the  minimum cost:
\cutequationup
{\small
\begin{align*}
    {\tau}^* = \argmin_{\tau \in \mathcal{T}({\bx}_0)}  f({\tau}, \mathcal{M}, \mathcal{O}, \mathcal{K}, \mathcal{V}; \bw) 
\end{align*}
}
with $f$ the scoring function, $\bw$  the learnable parameters of our models, $\mathcal{M}$  the map layers, $\mathcal{O}, \mathcal{K}, \mathcal{V}$ the occupancy and motion mode-probability and vector layers respectively, and $\mathcal{T}({\bx}_0)$ represents the possible trajectories which are generated conditioned on the current state of the SDV $\bx_0$. %

\cutsubsubsectionup
\subsubsection{Trajectory Sampling}
\cutsubsubsectiondown
The lane centers and topology are  strong priors to construct the potential trajectories to be  executed by the SDV. When an HD map is available, the lane geometry can be exploited to guide the trajectory sampling process. A popular approach, for example, is to sample trajectories in Frenet-frame of the goal lane-center,   limiting the samples to motions that do not deviate much from the desired lane \cite{werling2010optimal, ajanovic2018search, sadat2019jointly, sadat2020perceive}.
However, in  mapless driving  we need to take a different approach as the HD map is not available. We thus use retrieval  from a large-scale dataset of real trajectories. This approach provides a large set of trajectories from expert demonstrations while avoiding random sampling or arbitrary choices of acceleration/steering profiles \cite{phan2019covernet,zeng2019neural}.  
We create a dataset of expert demonstrations by binning based on the SDV initial state, clustering the trajectories of each bin, and using the cluster prototypes for efficiency.
During online motion planning, we retrieve the trajectories of the bin specified by $(v_\bx, a_\bx, \kappa_\bx)$ with $\bx$ the current state of the SDV. However, the retrieved trajectories may have marginally different initial velocity and steering angle than the SDV. Hence, instead of directly using those trajectories, we use the acceleration and steering rate profiles, $(a, \dot{\kappa})_t, t=0, ..., T$, to rollout a bicycle model \cite{polack2017kinematic}
 starting from the initial SDV state. This process generates trajectories with continuous velocity and steering. This is in contrast to the simplistic approach of, e.g., \cite{philion2020lift} where a fixed set of template trajectories is used, ignoring the initial state of SDV.

\cutsubsubsectionup
\subsubsection{Route Prediction}
\cutsubsubsectiondown
When HD maps are available, the input route is typically given in the form of a sequence of lanes that the SDV should follow. In mapless driving however, this is not possible. Instead,  we assume we are given a driving command as a tuple $c = (a, d)$, where $a \in \{ \textit{keep lane, turn left, turn right}\}$ is a discrete high-level action, and $d$ an an approximate longitudinal distance to the action. This information is similar to what an off-the-shelf GPS-based navigation system provides to human drivers.
To simulate GPS errors, we randomly sample noise from a zero-mean Gaussian with 5m standard deviation. 
We model the route as a collection of Bernoulli random variables, one for each grid cell in BEV.
Given the driving command $c$ and the predicted map $\mathcal{M}$, a routing network predicts a dense probability map $\mathcal{R}$ in BEV.
The routing network is composed of 3 CNNs that act like a switch for the possible high-level actions $a$. Note that only the one corresponding to the given driving command will be run at inference. Together with the predicted map layers, we "rasterize" the longitudinal distance $d$ to the action as an additional channel (i.e., repeated spatially), and leverage CoordConv \cite{liu2018intriguing} to break the translation invariance of CNNs.
\cutsubsubsectionup
\subsubsection{Trajectory Scoring}
\cutsubsubsectiondown
\label{sec:scoring}
We use a linear  combination of the  following cost functions to score the sampled trajectories. More detailed explanations about the individual costs can be found in the \supp{}.
\cutparagraphup
\paragraph{Routing and Driving on Roads:} 
\cutparagraphdown
In order to encourage the SDV to perform the high-level  command, we use a scoring function that encourages trajectories that travel a larger distance  in regions with high probability in $\mathcal{R}$. 
We use the following score function:
\cutequationup
{\small
\begin{align*}
    f_r(\tau, \mathcal{R})= - m(\tau) \min_{i \in m(\tau)} \mathcal{R}_i
\end{align*}
}
where $m(\tau)$ is the BEV grid-cells that overlap with SDV polygon in trajectory $\tau$. This score function makes sure the SDV stays on the route and is only rewarded when moving within the route. 
We introduce an additional cost-to-go that considers the predicted route beyond the planning horizon. This is important when there is a turn at the end of the horizon and the SDV velocity is high. Specifically, we compute the average value of $1-\mathcal{R}_j$ for all BEV grid-cells $j$ that have overlap with SDV beyond the trajectory horizon, assuming that the SDV maintains constant velocity and heading.

The SDV needs to always stay close to the center of the reachable lanes while on the road. Hence we use the predicted reachable lanes distance transform $\mathcal{M}^D$ to  penalize  distant trajectory points. 
In order to promote cautious behavior when there is high uncertainty in $\mathcal{M}^D$ and $\mathcal{M}^\theta$, we use a cost function that is the product of the SDV velocity and the standard deviation of the probability distributions of cells overlapping with SDV in $\mathcal{M}^D$ and $\mathcal{M}^\theta$. This promotes slow maneuver in the presence of map uncertainty.

The SDV is also required to  stay on the road and avoid encroaching onto the side-walks or the curb. Hence, we use the predicted drivable area $\mathcal{M}^A$ to penalize trajectories that go off the road:
\cutequationup
{\small
\begin{align*}
    f_a(\bx, \mathcal{M})= \max_{i \in m(\bx)} [1-P(\mathcal{M}^A_{i})]
\end{align*}
}
where $m(\bx)$ is the set of BEV grid-cells that overlap with SDV at trajectory point $\bx$. Similarly, the SDV needs to avoid junctions with red-traffic lights. Hence. we  use the predicted junction probability map $\mathcal{M}^J$ to penalize maneuvers that violate red-traffic light, similar to the routing cost. 

\cutparagraphup
\paragraph{Safety:}
\cutparagraphdown
The predicted occupancy layers and motion predictions are used to score the trajectory samples with respect to safety. We penalize trajectories where the SDV overlaps  occupied regions. 
For each trajectory point, we use the  BEV grid-cell with maximum probability among all the grid-cells that overlap with SDV polygon and use this probability directly as collision cost. The max operator ensures that the worst-case occupancy is considered over the region SDV occupies. 

The above objective promotes trajectories that do not overlap with occupied regions. However, the SDV needs to also maintain a safe distance from objects that are in the direction of SDV motion. This headway distance is a function of the relative speed  of the SDV wrt the other objects. 
To compute this cost for each trajectory point $\bx$, we retrieve all the BEV grid-cells in front of the SDV at $\bx$ and measure the violation of safety distance if the object  at each of those grid-cells stops with hard deceleration, and SDV with state $\bx_t$ reacts with a comfortable deceleration.
\cutparagraphup
\paragraph{Comfort:}
\cutparagraphdown
We also penalize jerk, lateral acceleration, curvature and its rate of change to promote comfortable driving.

\begin{table*}[t]
    \small
    \vspace{-20pt}
	\centering
	\begin{threeparttable}
        \begin{tabularx}{\textwidth}{
            l | r | r | r | s s s s s | r r  %
        }
		    \toprule
                Model & 
                Success & OffRoute & L2 & \multicolumn{5}{c|}{Progress per event $(m)$ $\uparrow$} & \multicolumn{2}{c}{Comfort} \\ %
                 & $(\%)$$\uparrow$ &  $(\%)$$\downarrow$ & $(m)$$\downarrow$ & any event &  collision & off-road  & off-route & oncoming & jerk$(\frac{m}{s^3})$ $\downarrow$ & lat.acc. $(\frac{m}{s^2})$$\downarrow$ \\ %
            \midrule
                IL &0.00 &99.39 & 39.10 &15.69 &44.49 &36.40 &30.28 &65.18 &98.99 &0.91 \\
                CIL &0.00 &99.39 & 35.53 &15.85 &38.50 &34.68 &35.64 &54.58 &52.88 &0.81 \\
                TC &12.80 &67.07 & 30.35 &51.17 &127.87 &288.07 &105.26 &329.90 &3.15 &0.25 \\
                NMP &22.56 &64.02 & 27.95 &69.83 &331.81 &721.74 &104.70 &1229.82 &3.04 &0.14 \\
                CNMP &21.34 &47.56 & 27.45 &74.85 &158.85 &646.49 &198.28 &543.32 &2.96 &0.26 \\
            \midrule
                \ourmodel{} &\textbf{74.39} &\textbf{14.63} & \textbf{12.95} &\textbf{218.40} &\textbf{1037.08} &\textbf{1136.49} &\textbf{409.34} &\textbf{1465.27} &\textbf{1.64} &\textbf{0.10} \\
	        \bottomrule
		\end{tabularx}
    \end{threeparttable}
    \cuthalfcaptionup
    \caption{
        \textbf{Closed-loop simulation results}
    }
	\label{table:closed_loop}
\end{table*}
\begin{table*}[t]
  \small
  \vspace{-5pt}
	\centering
	\begin{threeparttable}
        \begin{tabularx}{\textwidth}{l|rr|rr|s|ssr|ss}
            \toprule
            Model & \multicolumn{2}{c|}{Collisions $(\%)$} & \multicolumn{2}{c|}{L2 $(m)$} & Progress$(m)$ & OffRoute$(\%)$ & OffRoad$(\%)$& Oncoming$(\%)$ & lat.acc.$(\frac{m}{s^2})$ & Jerk $(\frac{m}{s^3})$ \\
            {} &            0-3s &   0-5s &    @3s &  @5s &         0-5s &             0-5s &            0-5s &                 0-5s &                 0-5s &         0-5s \\
            \midrule
            IL        &           2.17 &  9.54 &   \textbf{1.36} & \textbf{3.77} &        23.62 &             5.05 &            4.46 &                 3.05 &                 \textbf{1.00} &         2.47 \\
            CIL       &           2.20 & 10.15 &   1.38 & 3.79 &        23.58 &             5.16 &            5.28 &                 3.64 &                 1.10 &         2.60 \\
            TC &           1.72 &  6.95 &   2.02 & 4.34 &        22.26 &             2.68 &            0.28 &                 0.62 &                 1.47 &         7.48 \\
            NMP                &           0.83 &  5.18 &   1.75 & 4.47 &        23.09 &             1.59 &           \textbf{0.00}  &                 0.21 &                 1.14 &         3.98 \\
            CNMP               &           1.03 &  5.45 &   1.62 & 4.02 &        22.99 &             \textbf{0.14} &            0.07 &        0.14         &                 1.28 &         3.97 \\
            \midrule
            \ourmodel{}       &           \textbf{0.21} &  \textbf{2.07} &   1.71 & 4.54 &    \textbf{25.15} &             0.15 &            0.42 &        \textbf{0.09}          &                 1.23 &    \textbf{1.88}         \\
            \bottomrule
		\end{tabularx}
    \end{threeparttable}
    \cuthalfcaptionup
    \caption{\textbf{Large-scale evaluation against expert demonstrations} }
    \cutcaptiondown
	\label{table:open_loop}
\end{table*}
\cutsubsectionup
\subsection{Learning} \label{sec:learning}
\cutsubsectiondown
We optimize our driving model  in two stages. We first train the online map, dynamic occupancy field, and routing. Once these are converged, in a second stage, we keep these parts frozen and train the planner weights for the linear combination of scoring functions.
We found this 2-stage training empirically more stable than training end-to-end.

\cutparagraphup
\paragraph{Online map:}
\cutparagraphdown
We train the online map using negative log-likelihood (NLL) under the data distribution. That means, Gaussian NLL for  reachable lanes distance transform $\mathcal{M}^D$, Von Mises NLL for  direction of traffic $\mathcal{M}^\theta$ and binary cross-entropy for drivable area $\mathcal{M}^A$ and junctions $\mathcal{M}^J$.

\cutparagraphup
\paragraph{Dynamic occupancy field:} 
\cutparagraphdown
To learn the occupancy $\mathcal{O}$ of dynamic objects at the current  and  future time stamps, we employ  cross entropy loss with hard negative mining 
to tackle the high imbalance in the data (i.e., the majority of the space is free). 
To learn the probabilistic motion field, the motion modes $\mathcal{K}$ are learned in an unsupervised fashion via a categorical cross-entropy, where the true mode is defined as the one which associated motion vector is closest to the ground-truth motion in $\ell_2$ distance. Then, only the associated motion vector from the true mode is trained via a Huber loss.
Note that because the occupancy at future time steps $t > 1$ is obtained by warping the initial occupancy iteratively with the motion field, the whole motion field receives supervision from the occupancy loss. This is important in practice.
\cutparagraphup
\paragraph{Routing:}
\cutparagraphdown
We train the route prediction with binary cross-entropy loss. To learn a better routing model, we leverage supervision for all possible commands given a scene, instead of just the command that the SDV followed in the observational data. This does not require additional human annotations, since we can extract all possible (command, route) pairs from the ground-truth HD map. %
\cutparagraphup
\paragraph{Scoring:}
\cutparagraphdown
Since selecting the minimum-cost trajectory within a discrete set is non-differentiable, we use the max-margin loss \cite{maxmarginplan, sadat2020perceive} to penalize trajectories that have small cost but differ from the human demonstration or are unsafe.

\begin{figure*}[t]
    \vspace{-20pt}
    \centering
    \begin{tabular} {c@{\hspace{.5em}}c@{\hspace{.5em}}c@{\hspace{.5em}}c@{\hspace{.5em}}c}
        {} & \textbf{Scenario 1 - Keep Lane} & \textbf{Scenario 2 - Turn Left} & \textbf{Scenario 3 - Turn Right} & \multirow{3}{*}{\includegraphics[width=0.075\linewidth]{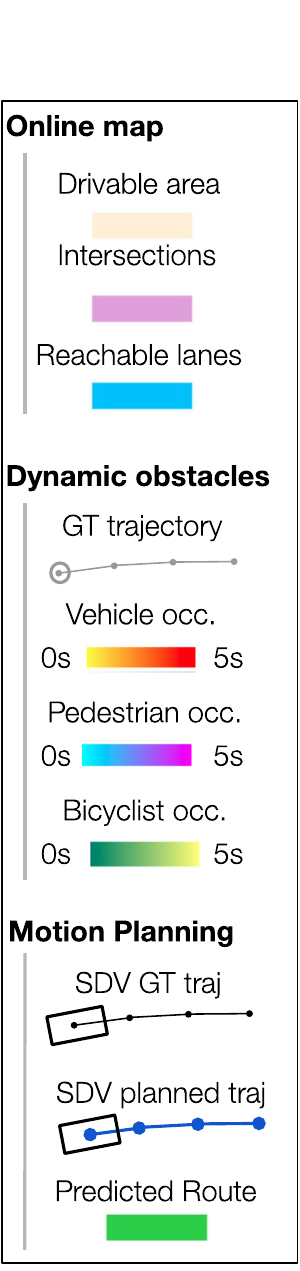}} \\
        \rotatebox[origin=c]{90}{\textbf{Map and Route}} &
        \raisebox{-0.5\height}{\includegraphics[width=0.26\linewidth, trim={3.0cm, 5.0cm, 1.0cm, 0.8cm}, clip]{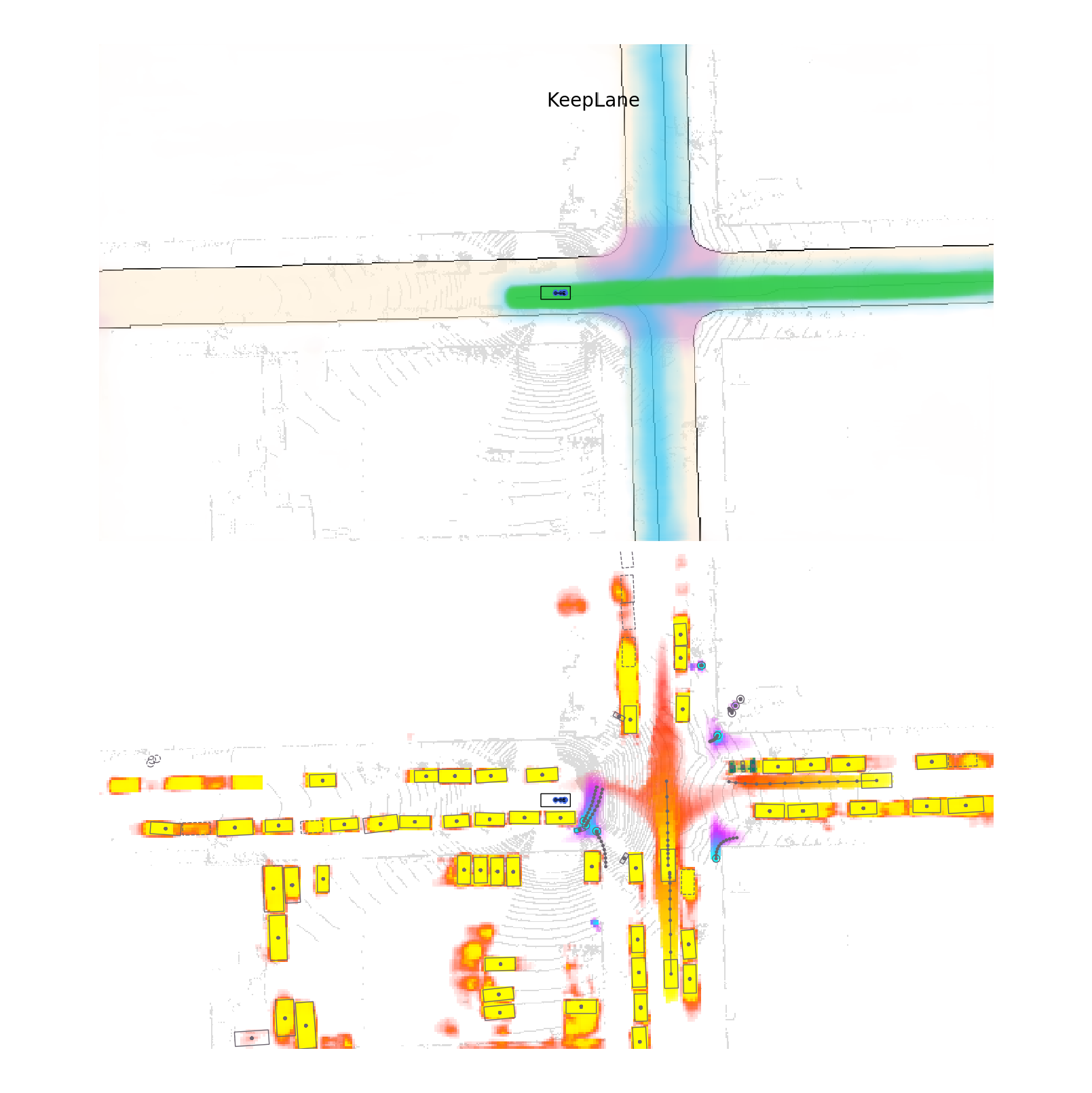}} &
        \raisebox{-0.5\height}{\includegraphics[width=0.26\linewidth, trim={3.0cm, 5.0cm, 1.0cm, 0.8cm}, clip]{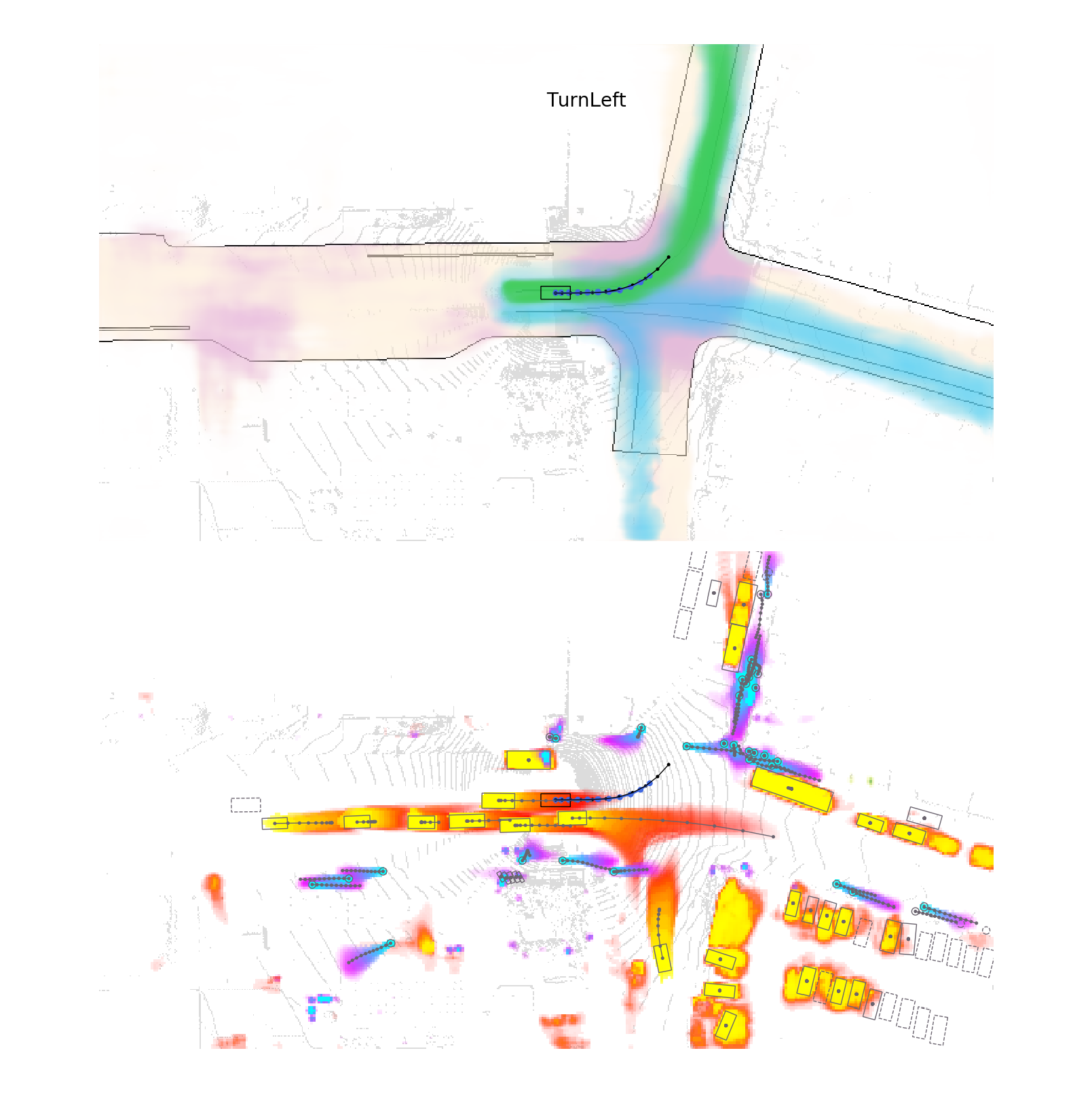}} &
        \raisebox{-0.5\height}{\includegraphics[width=0.26\linewidth, trim={3.0cm, 4.7cm, 1.0cm, 1.1cm}, clip]{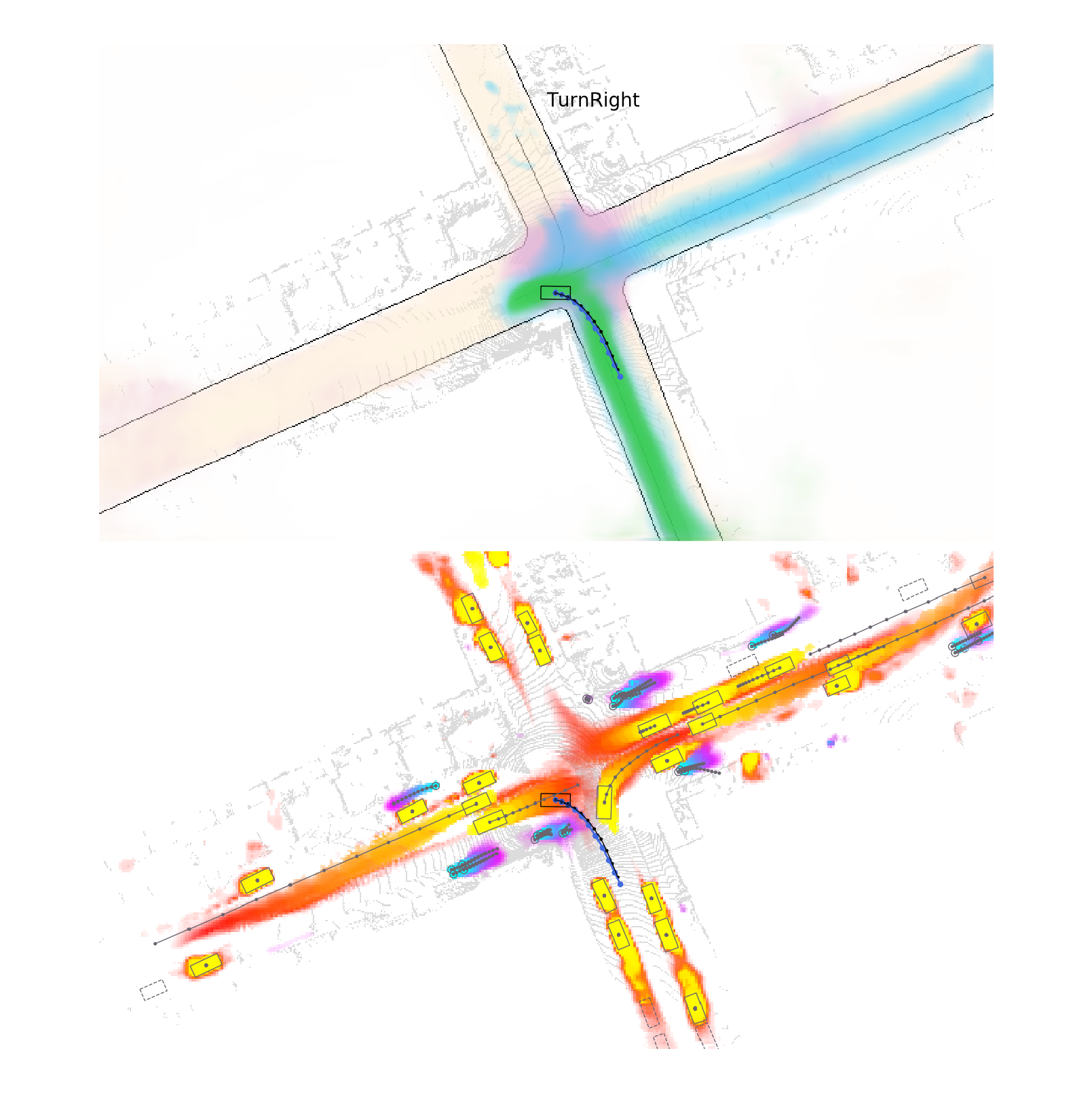}} & {}\\
        \rotatebox[origin=c]{90}{\textbf{Occupancy}} &
        \raisebox{-0.5\height}{\includegraphics[width=0.26\linewidth, trim={3.0cm, 1.0cm, 1.0cm, 4.8cm}, clip]{imgs/qualitative/0614b72c-c177-13d6-7289-f3d8669d9a0c_050.png}} &
        \raisebox{-0.5\height}{\includegraphics[width=0.26\linewidth, trim={3.0cm, 1.0cm, 1.0cm, 4.8cm}, clip]{imgs/qualitative/e8949091-59f1-db69-ed64-9d185ffdff7e_100.png}} &
        \raisebox{-0.5\height}{\includegraphics[width=0.26\linewidth, trim={3.0cm, 0.7cm, 1.0cm, 5.1cm}, clip]{imgs/qualitative/0e14076e-28b1-7dcb-aad5-260a4b58f0ed_150.png}} \vspace{0.1cm} & {} \\
    \end{tabular}
    \cuthalfcaptionup
    \caption{\textbf{Qualitative results.} We show our predicted scene representations and motion plan for different high-level actions.}
    \cutcaptiondown
    \label{fig:qualitative}
\end{figure*}

\cutsectionup
\section{Experimental Evaluation}
\cutsectiondown
In this section we first describe our experimental setup, and then present quantitative results in both closed-loop and open-loop.
Closed-loop evaluations are of critical importance since as the execution unrolls, the SDV finds itself in states induced by its own previous motion plans, and thus it is much more challenging than open-loop and closer to the real task of driving.
We defer the ablations of several components from our model to the \supp{}.

\cutparagraphup
\paragraph{Dataset:}
\cutparagraphdown
We train our models using our large-scale dataset \ourdataset{} that includes challenging scenarios where the operators are instructed to drive smoothly and in a safe manner. It contains 5000 scenarios for training, 500 for  validation and 1000 for the test set. Each scenario is 25 seconds. Compared to KITTI\cite{geiger2012we}, \ourdataset{} has 33x more hours of driving. Note that the train/validation/test splits are geographically non-overlapping which is crucial to evaluate generalization.
 
\cutparagraphup
\paragraph{Baselines:}
\cutparagraphdown
We compare against many SOTA approaches.
\textit{Imitation Learning (IL)}, where the future positions of the SDV are predicted directly from the scene context features, and is trained using L2 loss. \textit{Conditional Imitation Learning (CIL)} \cite{codevilla2018end}, which is similar to \textit{IL} but the trajectory is conditioned on the driving command.
\textit{Neural Motion Planner (NMP)} \cite{zeng2019neural}, where a planning cost-volume as well as detection and prediction are predicted in a multi-task fashion from the scene context features, and \textit{Trajectory Classification (TC)} \cite{philion2020lift}, where a cost-volume is predicted similar to NMP, but the trajectory cost is used to create a probability distribution over the trajectories and is trained by optimizing for the likelihood of the expert trajectory.
Finally, we extend NMP to consider the high-level command by learning a separate costing network for each discrete action (\textit{CNMP}).

\cutparagraphup
\paragraph{Closed-loop Simulation Results:}
\cutparagraphdown
Our simulated environment leverages a state-of-the-art LiDAR simulator \cite{manivasagam2020lidarsim} to recreate a virtual world from previously collected real static environments and a large-scale bank of diverse actors.
We use a set of 164 curated scenarios (18 seconds each) that are particularly challenging and require complex decision making and motion planning.
The simulation starts by replaying the motion of the actors as happened during the real-world capture. In case the scenario diverges from the original one due to SDV actions (e.g.,~SDV moving slower), the affected actors (e.g.,~rear vehicles) switch to the \textit{Intelligent Driver Model} \cite{Treiber_2000} for the rest of the simulation in order to be reactive.
We stop the simulation if the execution diverges too far from the commanded route.
A scenario is a success iff there are no events, i.e., the SDV does not collide with other actors, follows the route, does not get out of the road nor into opposite traffic. We report the \textit{Success rate}.
Because the goal of an SDV is to reach a goal by following the driving commands, we report \textit{Off-route (\%)}, which measures the percentage of scenarios the SDV goes outside the route.
Since all simulated scenarios are initialized from a real log, we measure the average \textit{L2} distance to the trajectory demonstrated by the expert driver.
Progress is measured by recording the meters traveled until an event happens. We summarize this in the metric \textit{meters per event}, and show a breakdown per event category.
Table~\ref{table:closed_loop} shows that our method clearly outperforms all the baselines across all metrics. 
\ourmodel{} achieves over 3x the success rate, diverges from the route a third of the times, imitates the human expert driver at least twice as close, and progresses 3x more per event than any baseline, while also being the most comfortable.

\cutparagraphup
\paragraph{Open-Loop Evaluation:}
\cutparagraphdown
We evaluate our method against human expert demonstrations on \ourdataset.
We measure the safety of the planner via the \textit{\% of collisions} with the ground-truth actors up to each trajectory time step.
\textit{Progress} measures how far the SDV advanced along the route for the 5s planning horizon, 
and \textit{L2} the distance to the human expert trajectory at different time steps.
To illustrate the map and route understanding, we compute the \textit{road violation rate}, \textit{oncoming traffic violation rate}, and \textit{route violation rate}.
Finally, \textit{jerk} and \textit{lateral acceleration} show how comfortable the produced trajectories are.
As shown in Table~\ref{table:open_loop} our \ourmodel{} model produces the safest trajectories that in turn achieve the most progress and are the most comfortable. In terms of imitation, IL and CIL outperform the rest since they are optimized for this metric, but are very unsafe. Our model achieves similar map-related metrics than the best performing baselines (NMP/CNMP) in open-loop. 
We want to stress the fact that these experiments are open-loop, and thus the SDV always plans from an expert state.
Because of this, it is very unusual to diverge from the route/road.
We consider this a secondary evaluation that does not reflect very well the actual performance when executing these plans, but include it for completeness since previous methods \cite{zeng2019neural,philion2020lift} benchmark this way. 
Comparing these results to closed-loop, we can see that \ourmodel{} is much more robust than the baselines to the distributional shift incurred by the SDV unrolling its own plans over time.

\cutparagraphup
\paragraph{Qualitative Results:}
\cutparagraphdown
Fig.~\ref{fig:qualitative} showcases the outputs from our model.
\textit{Scenario 1} shows the predictions when our model is commanded to keep straight at the intersection. Our model recognizes and accurately predicts the future motion of pedestrians near the SDV that just came out of occlusion, and plans a safe stop  accordingly. Moreover, we can appreciate the high expressivity of our dynamic occupancy field at the bottom, which can capture highly multimodal behaviors such as the 3 modes of the vehicle heading north at the intersection.
\textit{Scenario 2} and \textit{Scenario 3} show how our model accurately predicts the route when given turning commands, as well as how planning can progress through crowded scenes similar to the human demonstrations.
\ifthenelse{\boolean{arxiv}}{
    See Appendix~\ref{sec:supp_qualitative} for visualizations of the retrieved trajectory samples from the motion planner together with their cost, as well as a comparison of closed-loop rollouts against the baselines.
}{
} 

\cutsectionup
\vspace{-5pt}
\section{Conclusion}
\cutsectiondown
In this paper, we have proposed an end-to-end model for mapless driving.
Importantly, our method produces probabilistic intermediate representations that are interpretable and ready-to-use as cost functions in our neural motion planner.
We showcased that our driving model is safer, more comfortable and progresses the most among SOTA approaches in a large-scale dataset.
Most importantly, when we evaluate our model in a closed-loop simulator without any additional training it is far more robust than the baselines, achieving very significant improvements across all metrics.

\pagebreak

{\small
\bibliographystyle{ieee_fullname}
\bibliography{egbib}
}

\clearpage
\appendix
{\noindent \Large \textbf{\Supp} \vspace{0.15cm}} \\

In this \supp{} we first explain our implementation details in depth, then showcase additional experiments to provide more insights from our model, and finally provide additional qualitative results. 

\ifthenelse{\boolean{arxiv}}{

}{
    Please check out our supplementary video for a short summary of our method as well as qualitative results with narrated comments.
}

\section{Implementation details}

\subsection{Architecture}
\paragraph{Backbone network:} \label{sec:backbone}

\begin{figure*}[h]
    \centering
    \includegraphics[width=0.9\textwidth]{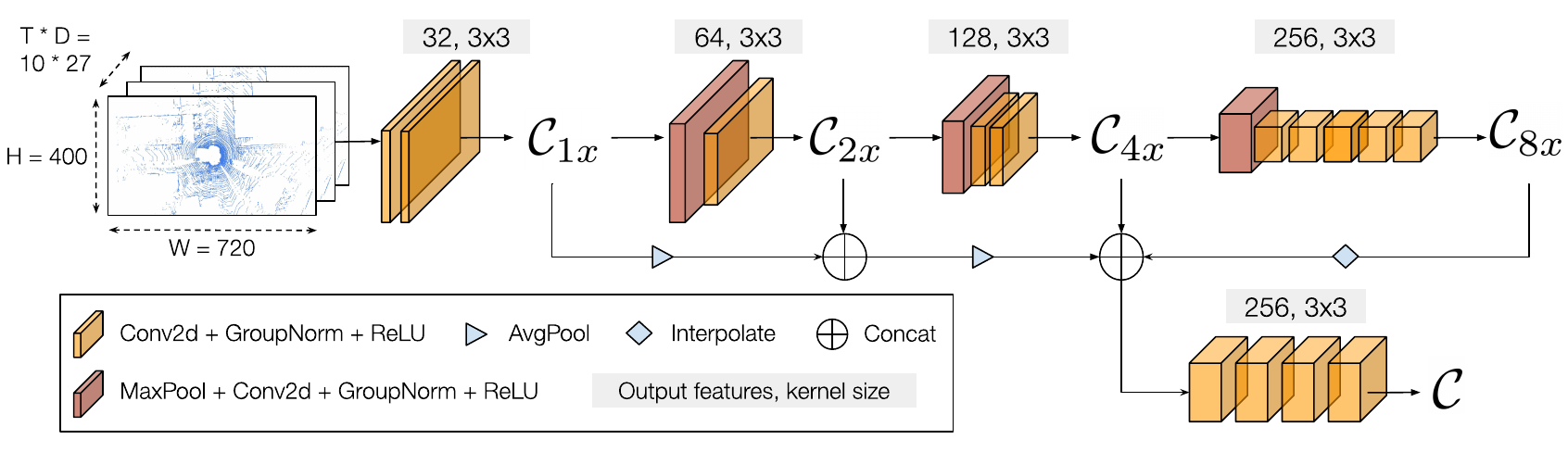}
    \caption{\textbf{Backbone Network}. The output features within one CNN block are fixed. All kernel strides and dilation are 1.} 
    \label{fig:backbone}
 \end{figure*}

Fig.~\ref{fig:backbone} shows an architecture diagram of the backbone network. We combine ideas from \cite{yang2018pixor,casas2018intentnet} to  build a multi-resolution backbone network that extracts geometric and semantic information from the past LiDAR sweeps and is able to aggregate them to reason about motion.
Our backbone is composed of 4 convolutional blocks, and a convolutional header. The number of output features and kernel size is indicated in Fig.~\ref{fig:backbone}. All convolutional layers use Group Normalization \cite{wu2018group} with 32 groups, and ReLU non-linearity. The scene context features after each residual block are $\mathcal{C}_{1x}, \mathcal{C}_{2x}, \mathcal{C}_{4x}, \mathcal{C}_{8x}$, where the subscript indicates the downsampling factor from the input in BEV. The features from the different blocks are then concatenated at 4x downsampling by max-pooling higher resolution ones $\mathcal{C}_{1x}, \mathcal{C}_{2x}$ and interpolating $\mathcal{C}_{8x}$. 
Finally, a residual block of 4 convolutional layers with no downsampling outputs the scene context $\mathcal{C}$. Since the LiDAR input is voxelized at 0.2 meters per pixel, $\mathcal{C}$ has a resolution of 0.8 meters per pixel.

\paragraph{Mapping architecture:}

\begin{figure*}[h]
    \centering
    \includegraphics[width=0.9\textwidth]{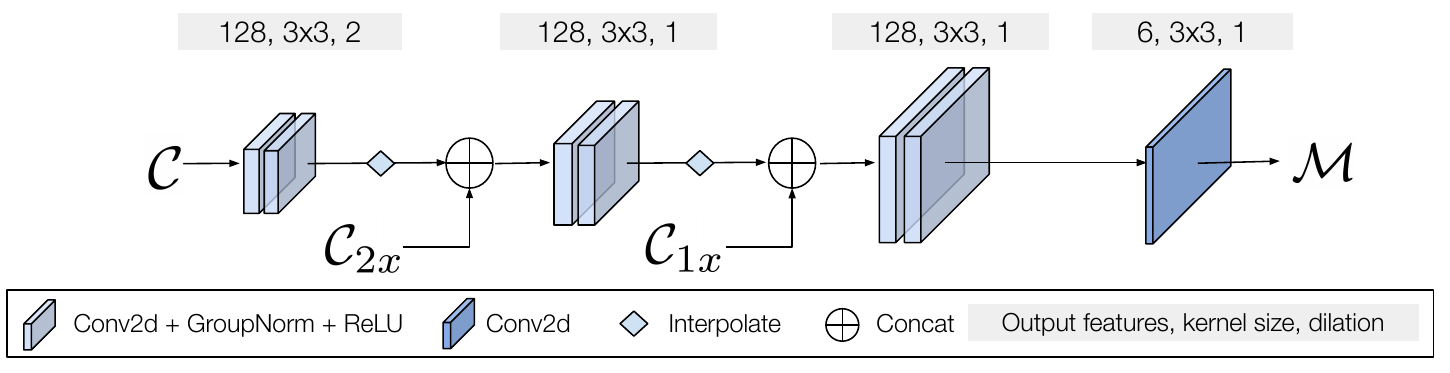}
    \vspace{-10pt}
    \caption{\textbf{Mapping Network}. The output features within one CNN block are fixed. All kernel strides are 1.} 
    \label{fig:mapping}
 \end{figure*}

In order to drive safely (e.g., narrow streets) we need a high resolution representation of our maps, and at the same time a big receptive field to help reduce uncertainty (e.g., around  occlusions). 
To achieve this efficiently, we employ a multi-resolution architecture as shown in Fig.~\ref{fig:mapping}. It takes as input multiple feature maps $\mathcal{C}_{1x}, \mathcal{C}_{2x}, \mathcal{C}$ from the backbone network (see Fig.~\ref{fig:backbone}), and outputs the 6 channels of the online map at the original input resolution of 0.2 m/pixel. As a reminder, the six channels are: 1 for drivable area score, 1 for intersection score, 2 for truncated unsigned distance reachable lane (mean and variance of Gaussian distribution), 2 for the angle to closest reachable lane segment (location and concentration of Von Mises distribution).

\paragraph{Perception and Prediction architecture:}

The different dynamic classes (i.e. vehicles, pedestrians and bicyclists) are processed by different networks since they have very different geometry as well as motion. For each class, a 2-layer CNN processes $\mathcal{C}$ and upsamples it to 0.4 m/pixel. This is the same resolution as $\mathcal{C}_{2x}$, which gets processed by another 2-layer CNN. 
Then the two feature maps get concatenated to form the dynamic  context. From this context, a 1-layer CNN outputs the current occupancy map, a 2-layer CNN the motion mode scores for all time steps, and another 2-layer CNN the motion vectors for all modes and future time steps. We employ dilation \cite{yu2015multi} to increase the receptive field  while keeping the number of parameters low in order to be able to predict motion for voxels that are far away from the original position of actors for the long temporal horizons. To infer the future occupancy, we simply warp the initial occupancy with the temporal motion field as explained in the main manuscript's Fig. 3.4., and further detailed in Section~\ref{sec:supp_occ}. In practice, we found that $K=3$ modes was expressive enough for the multi-modality of the temporal motion field. In all our experiments, $T=11$ since we predict the future 5 seconds at 0.5-second intervals.

\begin{figure*}[h]
    \centering
    \vspace{-15pt}
    \includegraphics[width=0.9\textwidth]{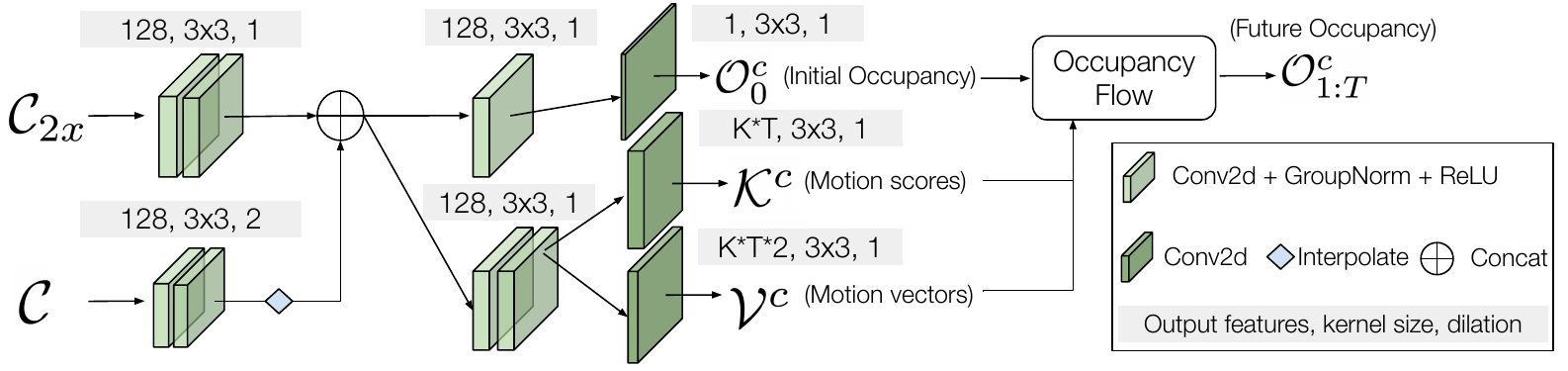}
    \caption{\textbf{Perception and Prediction Network}. It outputs the initial occupancy and current and future motion. The output features within one CNN block are fixed. All kernel strides are 1.} 
    \label{fig:dynamic_occupancy}
 \end{figure*}

\paragraph{Routing architecture:}

\begin{figure*}[h]
    \centering
    \includegraphics[width=0.9\textwidth]{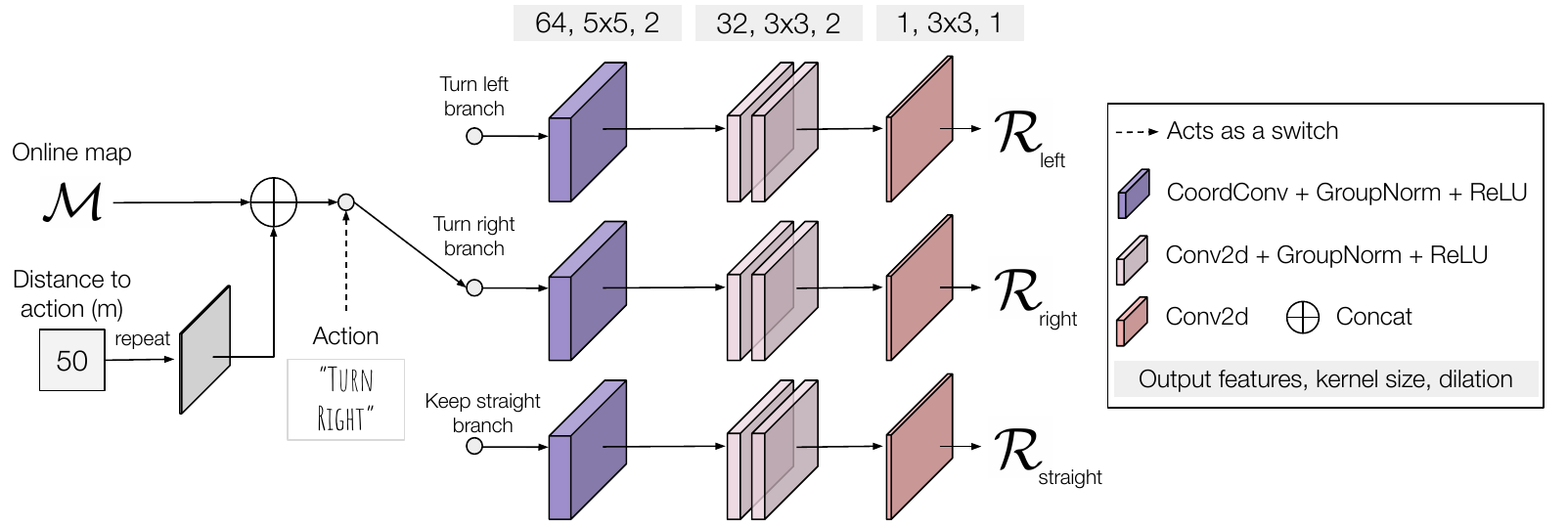}
    \caption{\textbf{Routing Network}. The output features within one CNN block are fixed. All kernel strides are 1. The architecture is the same for the 3 branches, but with different learnable parameters.} 
    \label{fig:routing}
 \end{figure*}

In order to drive towards a goal, we would like to follow the driving commands. To do so, we predict a route spatial map, where each cell represents the probability that driving to it from the current location is aligned with the driving command. The architecture for the network that predicts such spatial map is detailed in \ref{fig:routing}. 
The high-level action in the command acts as a switch between 3 instantiations of the same network architecture (i.e., one for turning right, one for turning left, one for going straight). The longitudinal distance to action is repeated spatially with the same resolution as the online map. Then, both are concatenated to form the input to a CNN that leverages Coordinate Convolutions (CoordConv) \cite{liu2018intriguing} in order to be able to reason about the distance to a particular grid cell from the SDV.

\subsection{Dynamic occupancy} \label{sec:supp_occ}

Here we provide a more detailed explanation of how our dynamic occupancy flow works than in the main manuscript.
We first define a \textit{flow event} from spatio-temporal grid cell $(t, i_1)$ to $(t+1, i_2)$ as the intersection of the event that the original grid cell is occupied, and that the motion field moves this occupancy from   
$(t, i_1)$ to $(t+1, i_2)$. Since we consider $K$ motion modes, the final flow event considers the union of those, effectively marginalizing over modes:
$$
\mathcal{F}_{(t, i_1) \to (t+1, i_2)} = \cup_k  \{\mathcal{O}^{t}_{i_1} \cap \mathcal{K}^t_{i_1}=k \cap \mathcal{V}^{t, k}_{i_1}=i_2 \}
$$
Given this flow event definition and the assumptions in our probabilistic model, we can obtain its probability:
{\small
\begin{align*}
p(\mathcal{F}^c_{(t, i_1) \to (t+1, i_2)}) = \sum_k p(\mathcal{O}^c_{t, i_1}) p(\mathcal{K}^c_{t,i_1}=k) p(\mathcal{V}^c_{t, i_1, k}=i_2)
\end{align*}
}

We can now calculate the future occupancy iteratively, starting from the occupancy predictions at $t=0$. 
Specifically, to get the occupancy that flows into cell $i$ at time $t+1$ from all cells $j$ at time $t$, we can simply compute the probability that no occupancy flow event occurs, and take its complement
\begin{align*}
    p(\mathcal{O}^c_{t+1, i}) &= p\big(\cup_{j}  \mathcal{F}_{(t, j) \to (t+1, i)} : \forall k, l\big) \\
    &= 1 - p\big(\cap_{j}  \mathcal{F}^{'}_{(t, j) \to (t+1, i)} : \forall k, l\big) \\
    &= 1 - \prod_{j}\big( 1 - p(\mathcal{F}^c_{(t, j) \to (t+1, i)})\big)
\end{align*}
where $\mathcal{F}^{'}_{(t, j) \to (t+1, i)}$ denotes the complement of $\mathcal{F}_{(t, j) \to (t+1, i)}$.

\subsection{Planning}
In this section we provide more details about trajectory retrieval and the coring functions.

\subsubsection{Trajectory Retreival}
\begin{figure*}[t]
    \centering
    \vspace{-30pt}
    \includegraphics[width=0.7\textwidth]{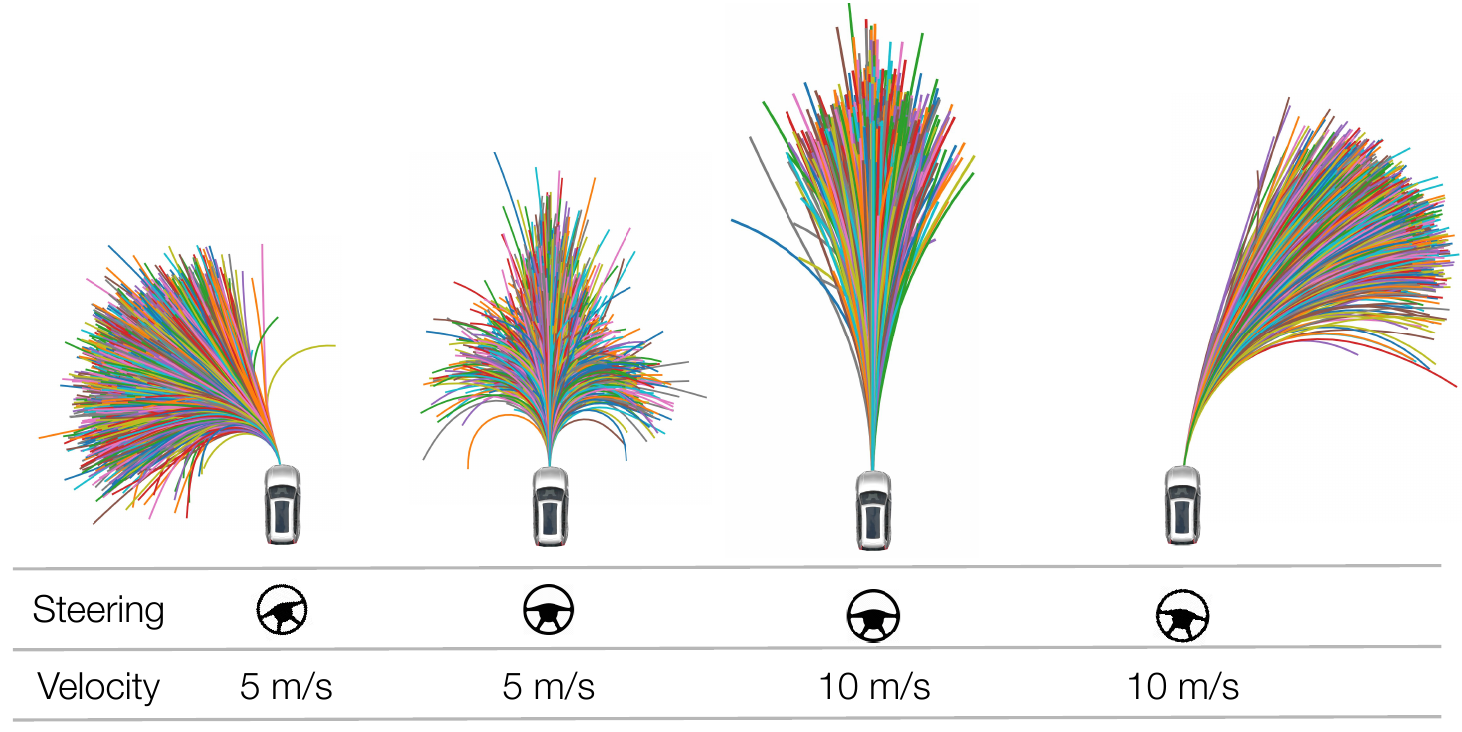}
    \caption{Sets of trajectories retrieved from the expert demonstrations. The initial state of the SDV is used to query trajectories that start with similar state (the initial acceleration is $0.0\frac{m}{s^2}$ in all the shown cases). } 
    \label{fig:traj_retrieval}
 \end{figure*}

We use $\sim$ 150 hours of manual driving data to create the dataset of expert trajectories. To group the trajectories into different bins, we use the initial velocity $v$, curvature $\kappa$, and acceleration $a$, with respective bin sizes of $2.0\frac{m}{s}$, $0.02\frac{1}{m}$, and $1.0\frac{m}{s^2}$. The trajectories in each bins are clustered into $3,000$ sets and the closest trajectory to each cluster prototype is kept. This creates a set fo diverse trajectories conditioned on the current state of SDV. Figure \ref{fig:traj_retrieval} shows example sets of trajectories retrieved base on the indicated initial state (initial acceleration is $0.0$ in all the cases). We can see how the set of trajectories are influenced by the initial state, resulting in kinematically-plausible trajectory sampling.

\subsubsection{Trajectory Scoring}
In the following, we provide more details about the cost function of the planner.
\paragraph{Reachable-lanes direction}
In addition to stay close to the lane centerlines, we promote trajectories that stay aligned with the direction of the lane. Towards this goal, we use the average difference of the trajectory point heading and the angles in $\mathcal{M}^\theta$ indexed at all BEV grid-cells that have overlap with SDV polygon. 
\begin{align*}
    \label{eq:direction}
    f_d(\bx, \mathcal{M}) = \mathop{\mathbb{E}}_{i \in m(\bx)} |\mathcal{M}^\theta_{i}-\bx_\theta|
\end{align*}
where $m(\bx)$ represents the spatial indices of BEV grid-cells of the map-layer prediction that overlaps with SDV polygon with state $\bx$.

\begin{table*}[t]
    \small
	\centering
	\begin{threeparttable}
        \begin{tabularx}{\textwidth}{
            l | r | r | r | s s s s s | r r  %
        }
		    \toprule
                Loss & 
                Success & OffRoute & L2 & \multicolumn{5}{c|}{Progress per event $(m)$ $\uparrow$} & \multicolumn{2}{c}{Comfort} \\ %
                 & $(\%)$$\uparrow$ &  $(\%)$$\downarrow$ & $(m)$$\downarrow$ & any event &  collision & off-road  & off-route & oncoming & jerk$(\frac{m}{s^3})$$\downarrow$ & lat.acc.$(\frac{m}{s^2})$$\downarrow$ \\ %
            \midrule
                Motion &68.90 &\textbf{11.59} & 13.10 &171.53 &545.59 &991.81 &\textbf{514.30} &1445.84 &1.73 & \textbf{0.10} \\
                + Occ. &\textbf{74.39} &14.63 & \textbf{12.95} &\textbf{218.40} &\textbf{1037.08} &\textbf{1136.49} &409.34 &\textbf{1465.27} &\textbf{1.64} &\textbf{0.10} \\
	        \bottomrule
		\end{tabularx}
    \end{threeparttable}
    \cuthalfcaptionup
    \caption{
        \textbf{Loss ablation for future dynamic occupancy} (closed-loop simulation results). Adding supervision to the warped occupancy drastically improves our model, compared to supervising only the motion field.
    }
    \cutcaptiondown
	\label{table:occ_ablation}
\end{table*}
\begin{table*}[t]
    \small
	\centering
	\begin{threeparttable}
        \begin{tabularx}{\textwidth}{
            l | r | r | r | s s s s s | r r  %
        }
		    \toprule
                Routing & 
                Success & OffRoute & L2 & \multicolumn{5}{c|}{Progress per event $(m)$ $\uparrow$} & \multicolumn{2}{c}{Comfort} \\ %
                input & $(\%)$$\uparrow$ &  $(\%)$$\downarrow$ & $(m)$$\downarrow$ & any event &  collision & off-road  & off-route & oncoming & jerk$(\frac{m}{s^3})$$\downarrow$ & lat.acc.$(\frac{m}{s^2})$$\downarrow$ \\ %
            \midrule
                None &46.95 &44.51 & 22.04 &93.44 &\textbf{1614.09} &1106.78 &116.52 &1402.32 &1.85 &\textbf{0.07} \\
                + action &67.68 &19.51 & 13.47 &166.58 &935.18 &\textbf{1255.28} &296.68 &1132.53 &1.75 &0.09 \\
                + dist. &\textbf{74.39} &\textbf{14.63} &\textbf{12.95} &\textbf{218.40} &1037.08 &1136.49 &\textbf{409.34} &\textbf{1465.27} &\textbf{1.64} &0.10 \\
	        \bottomrule
		\end{tabularx}
    \end{threeparttable}
    \cuthalfcaptionup
    \caption{
        \textbf{Route ablation} (closed-loop simulation results). Naturally, adding a discrete route command (action) allows the SDV to better progress towards the goal. Moreover, complementing it with a noisy longitudinal distance to action is helpful. 
    }
    \cutcaptiondown
	\label{table:route_ablation}
\end{table*}
\begin{table*}[t]
    \small
	\centering
	\begin{threeparttable}
        \begin{tabularx}{\textwidth}{
            l | r | r | r | s s s s s | r r  %
        }
		    \toprule
                Map \& & 
                Success & OffRoute & L2 & \multicolumn{5}{c|}{Progress per event $(m)$ $\uparrow$} & \multicolumn{2}{c}{Comfort} \\ %
                route & $(\%)$$\uparrow$ &  $(\%)$$\downarrow$ & $(m)$$\downarrow$ & any event &  collision & off-road  & off-route & oncoming & jerk$(\frac{m}{s^3})$$\downarrow$ & lat.acc.$(\frac{m}{s^2})$$\downarrow$ \\ %
            \midrule
                GT & 84.80 & 0.0  & 9.19 & 415.41 & 684.61 & 1822.48 & $\infty$  & 3689.89 &1.53 & 0.15 \\
                Pred. & 74.39 &14.63 & 12.95 &218.40  &1037.08 &1136.49  &409.34     &1465.27  &1.64 &0.10 \\
	        \bottomrule
		\end{tabularx}
    \end{threeparttable}
    \cuthalfcaptionup
    \caption{
        \textbf{Upper bound performance} in closed-loop simulation when using an HD map to feed the motion planner the true online map and route.
    }    
    \cutcaptiondown

	\label{table:supp_upper_bound}
\end{table*}

\paragraph{Lane uncertainty}
In order to promote cautious behavior when there is high uncertainty in $\mathcal{M}^D$ and $\mathcal{M}^\theta$, we use a cost function that is the product of the SDV velocity and the standard deviation of the probability distributions of cells overlapping with SDV in $\mathcal{M}^D$ and $\mathcal{M}^\theta$. This promotes slow maneuver in the presence of map uncertainty:
\begin{align*}
    f_d(\bx, \mathcal{M}^\theta, \mathcal{M}^D) = \sum_{i \in m(\bx)} \bx_v(\sigma^D_i + \frac{1}{k^\theta_i})
\end{align*}
Here $\sigma^D_i$ denotes the standard deviation of the Gaussian distribution representing distance to closest reachable lane center, and $k^\theta_i$ is the concentration parameter of the von Mises distribution representing lane direction.

\paragraph{Occupancy}
Given the state of the ego car $\bx$ for a single time-step, we use the following cost function to penalize trajectories that overlap with occupied regions:
\begin{align*}
    f_o(\bx_t, \mathcal{O})=\sum_c \max_{i \in m(\bx_t)} P(\mathcal{O}^c_{t,i})
\end{align*}
where $m(\bx_t)$ represents the BEV grid-cells, with semantic class $c$, that have overlap with the polygon of SDV with state $\bx_t$. 

\paragraph{Headway}
For the headway cost, we retrieve the set of BEV grid-cells $m(\bx_t)$ that are within 20m in front of the SDV at time $t$. The headway cost is then computed over this set as follows:
\begin{align*}
    f_h(\bx_t, \mathcal{O}, \mathcal{K}, \mathcal{V})=\sum_{i \in m(\bx_t)} P(\mathcal{O}_{t,i}) \mathop{\mathbb{E}}_{P(\mathcal{K}_{t,i})} [h(\bx_t, \mathcal{V}_{t,i})]
\end{align*}
where the expectation is over the different motion modes. The function $h(\bx_t, \mathcal{V}_{t, i})$ measures the violation of safety distance if the object  at spatial index $i$ with speed $\mathcal{V}_{t, i}$ stops with hard deceleration, and SDV with state $\bx_t$ reacts with a comfortable deceleration.

\subsection{Training details}
\paragraph{Multi-task learning (1st stage):}
In the first stage we train the backbone, mapping, perception and prediction as well as routing networks.
To do so, we linearly combine the mapping loss $\mathcal{L}_\mathcal{M}$, occupancy loss $\mathcal{L}_\mathcal{O}$, motion loss $\mathcal{L}_{\mathcal{K}, \mathcal{V}}$, and routing loss $\mathcal{L}_\mathcal{R}$ described in the main manuscript.
$$
\mathcal{L}_{\text{stage 1}} = \mathcal{L}_\mathcal{O} + \lambda_{\mathcal{K}, \mathcal{V}} \mathcal{L}_{\mathcal{K}, \mathcal{V}} + \lambda_\mathcal{M} \mathcal{L}_\mathcal{M} + \lambda_\mathcal{R} \mathcal{L}_\mathcal{R}
$$
where $\lambda_{\mathcal{K}, \mathcal{V}}=0.1, \lambda_{\mathcal{M}}=0.5, \lambda_\mathcal{R}=2.0$ are hyperparameters that were found to work well in practice in our validation set. Within these task losses, all losses are summed with equal weight (e.g., for mapping, the negative-log likelihood for each map element has the same weight).
\paragraph{Trajectory Scoring (2nd stage):} 
Since selecting the minimum-cost trajectory within a discrete set is non-differentiable, we use the max-margin loss \cite{maxmarginplan, sadat2020perceive} to penalize trajectories that have small cost but differ from the human demonstration or are unsafe. 
Let $\tau_h$ be the expert demonstration for a given example. 
The max-margin loss  encourages the human  driving trajectory to have smaller cost $f$ than other trajectories. 
{\small
\begin{align*}
\mLM = \max\limits_{\tau} \bigg[&~ f_{r}(\tau_{h})-f_{r}(\tau) +l_\text{im} + \sum_{t}\big[f^{t}_{o}(\tau_{h})-f^{t}_{o}(\tau) +l^{t}_\text{o} \big]_+ \bigg]_+
\end{align*}
}
where $f^t_o$ is the occupancy cost function at time step $t$, $f_{r}$ are the rest of the planning subcosts, and $[]_+$ represents the ReLU function. Note that in above, we dropped the other inputs to the cost functions for brevity.
The imitation task-loss $l_\text{im}$ measures  the $\ell_1$ distance  between trajectory $\tau$ and the ground-truth for the entire horizon, and the safety task-loss $l^t_\text{o}$ accounts for collisions and their severity at each trajectory step. By imposing the task-loss per time-step and separate from the other non-safety subcosts, we make sure the margin in cost is achieved when the trajectory is in collision (high $l^t_\text{o}$) irrespective of other costs at other time-steps.

\section{Additional Experiments}

\subsection{Dynamic occupancy loss ablation}

Our proposed model predicts multi-modal motion vectors and their categorical distribution for each spatio-temporal BEV grid cell, which is then used to flow the initial predicted occupancy to the future steps.
The most direct way of supervising the predicted elements, i.e., initial occupancy and temporal motion fields, is to have a separate loss for each as they both have supervision from the labels, similar to the loss in \cite{wu2020motionnet}.
However, this has 2 main drawbacks. 
First, this training strategy does not optimize future occupancy, which is critical for safety in motion planning. Because the future occupancy is obtained as a resulting of flowing (or warping) the initial occupancy with the temporal motion field, small errors in motion can accumulate over time.
Second, only the motion vector that is closest to the ground-truth gets supervision, leaving the rest of the distribution over possible motions unsupervised.
We implement this strategy and ablate it against our training, where we also apply supervision to the future occupancy after flow, thus optimizing for what we care about for safe driving, and also giving signal to the full motion distribution. The results shown in Table~\ref{fig:dynamic_occupancy} show a clear advantage of our training strategy. We would like to highlight that this improvement in the loss makes the model roughly twice as safe, as indicated by the progress per collision metric. While it is true that we suffer a minor regression in route following probably to avoid some collisions by deviating laterally, safety takes precedence.

\subsection{Routing ablation}

In order to show the importance of route prediction, we consider (i) no route prediction, (ii) predicting the route map based only on the input discrete high-level action coming from the command through switching between different NNs, (iii) also taking into account the approximate longitudinal distance to action as explained in Fig.~\ref{fig:routing}. 
Our results in Table~\ref{table:route_ablation} show several interesting aspects. First, we see that only by using the discrete high-level action (Action only), which is the same information as the CIL and CNMP baselines use, our model is able to achieve a much better route following than those. In particular, while the best baseline, CNMP, went 47.56\% of the times off-route, our \textit{MP3 - Action only} ablation only goes out of route 19.51\% of the times.
Finally, we see that adding the approximate distance to action and using CoordConv \cite{liu2018intriguing} such that the routing network can reason about the distance from SDV further reduces the out-of-route events to 14.63\%.
We note that the decrease in progress per collision we observe when adding the route is due to the fact that the base model is not constrained to the route, and therefore can swerve outside the route without cost to avoid collisions. In that case, our metrics only record an off-route event for the breakdown.
However, doing this consistently would make any travel very frustrating since it would make it impossible to reach the destination in time, and thus this is not a reasonable option.

\subsection{Upper bound with access to HD map}
Table~\ref{table:supp_upper_bound} showcases the results of our model when feeding the motion planner with the ground-truth online map and route map, instead of the predicted ones. For this experiment we only retrain the motion planner weights (i.e. the second stage in the training described in Sec.~\ref{sec:learning}).
We note that the planner is still not receiving a perfect representation of the scene, as it needs to infer the dynamic occupancy from sensor data.
This experiment shows several interesting aspects. 
We can see that with a perfect map and route prediction, the proposed motion planner (including the retrieval-based trajectory sampler) always follows the route. 
This experiment also confirms what we have observed in the other ablations: sometimes diverging from the route is an easier way to avoid collisions than a change in the speed profile, as shown by the progress per collision metric. 
Finally, this experiment confirms that the proposed representations for the online map and the route, as well as the motion planning costs are adequate, and motivates future work to improve the prediction of the static part of the environment.

\section{Qualitative results} \label{sec:supp_qualitative}
\subsection{MP3 outputs in closed-loop simulation}
\begin{figure*}[t]
    \vspace{-10pt}
    \centering
    \begin{tabular} {c}
        \includegraphics[width=0.8\linewidth]{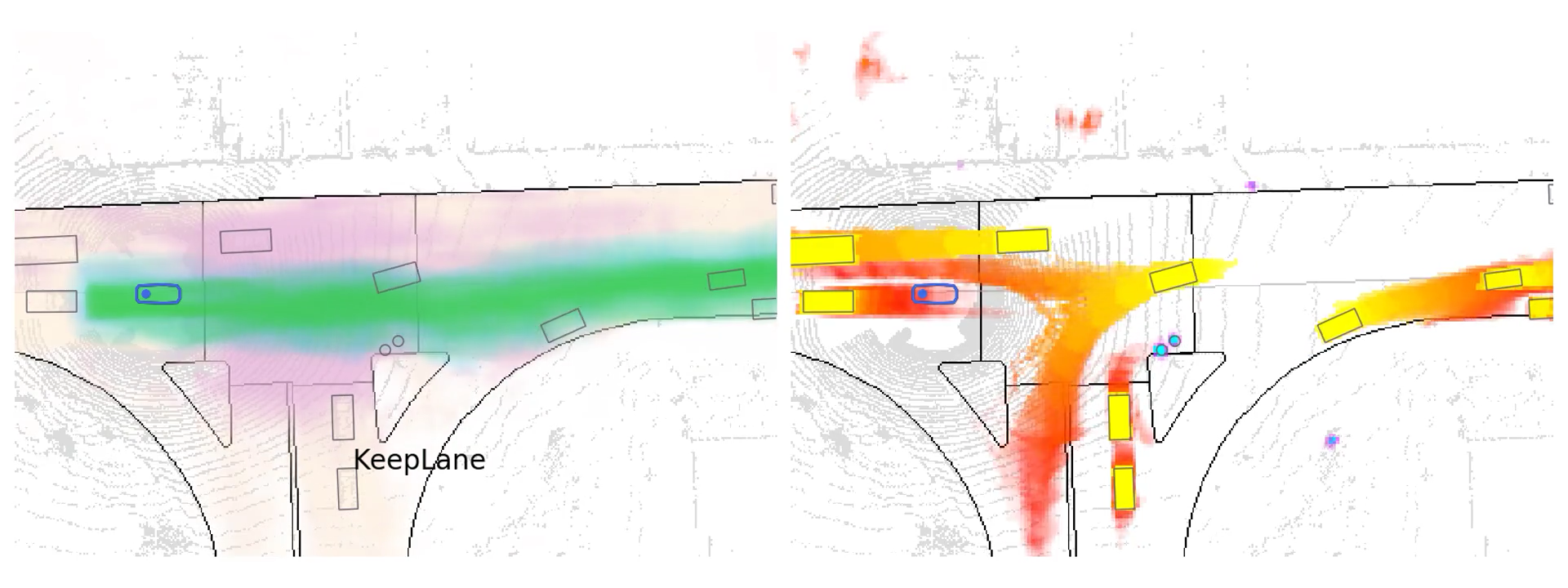} \\
        \includegraphics[width=0.8\linewidth]{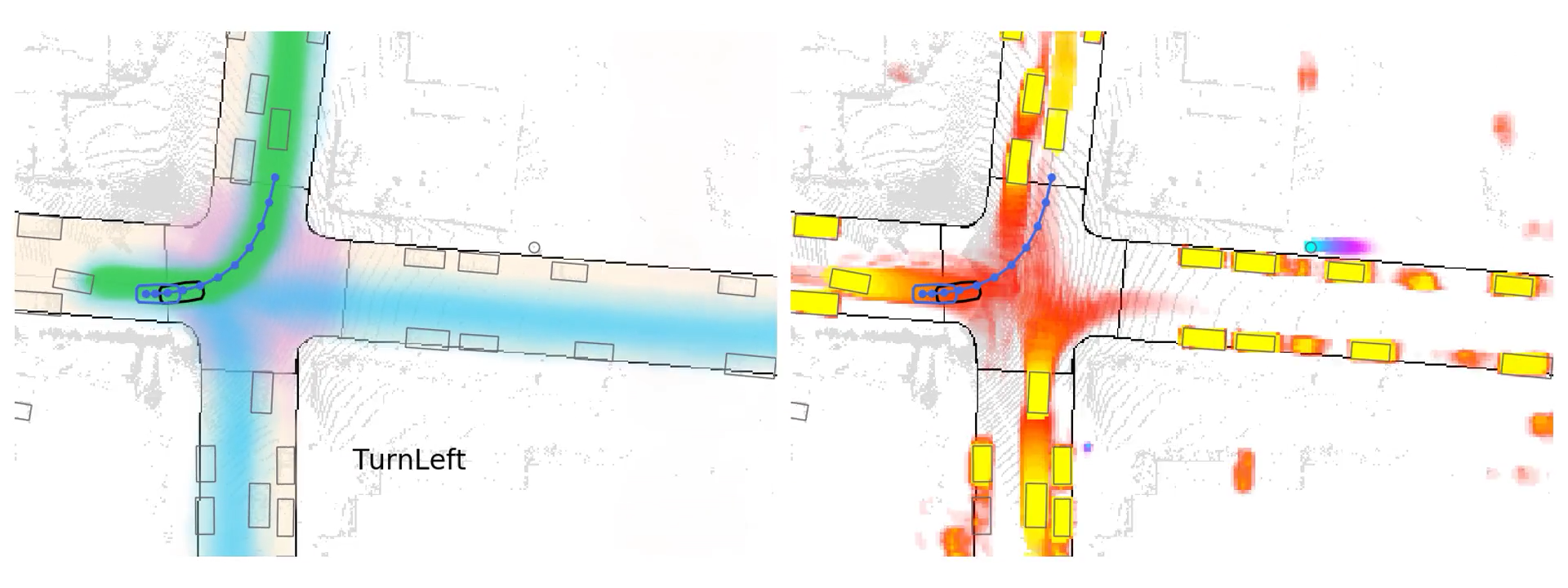} \\
        \includegraphics[width=0.8\linewidth]{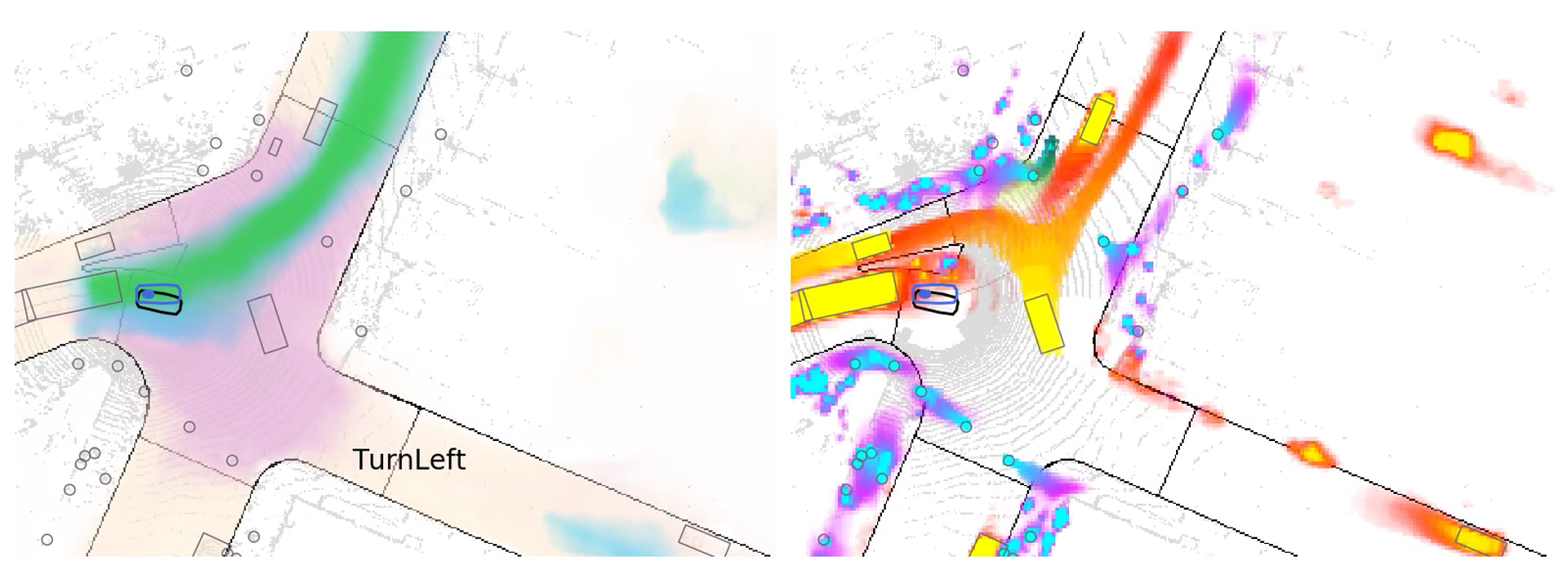} \\
        \includegraphics[width=0.8\linewidth]{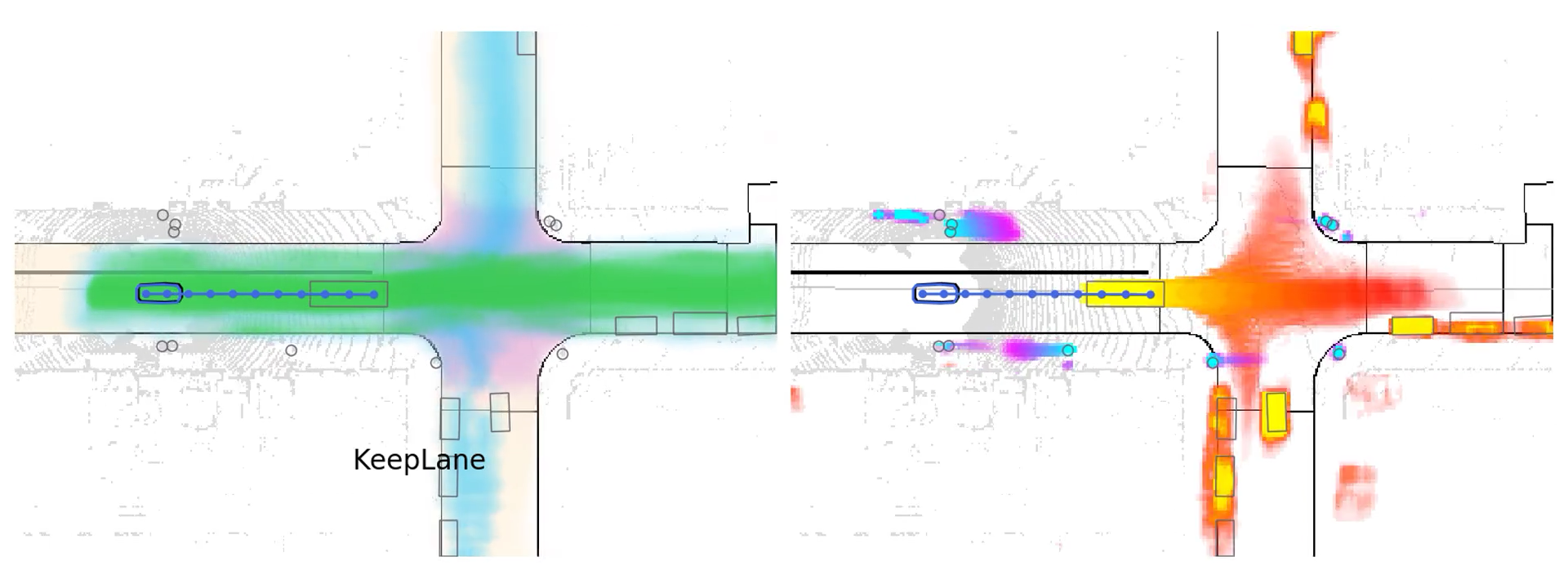} \\
    \end{tabular}
    \vspace{-10pt}
    \caption{\textbf{Qualitative results of MP3 in closed-loop.} We show our predicted scene representations and motion plan in different interesting scenarios. The black bounding box represents the state of the expert driver at that point, and we can see it's very close to the planner state (in blue)}
    \label{fig:qualitative_supp1}
\end{figure*}

\begin{figure*}[t]
    \vspace{-10pt}
    \centering
    \begin{tabular} {c}
        \includegraphics[width=0.8\linewidth]{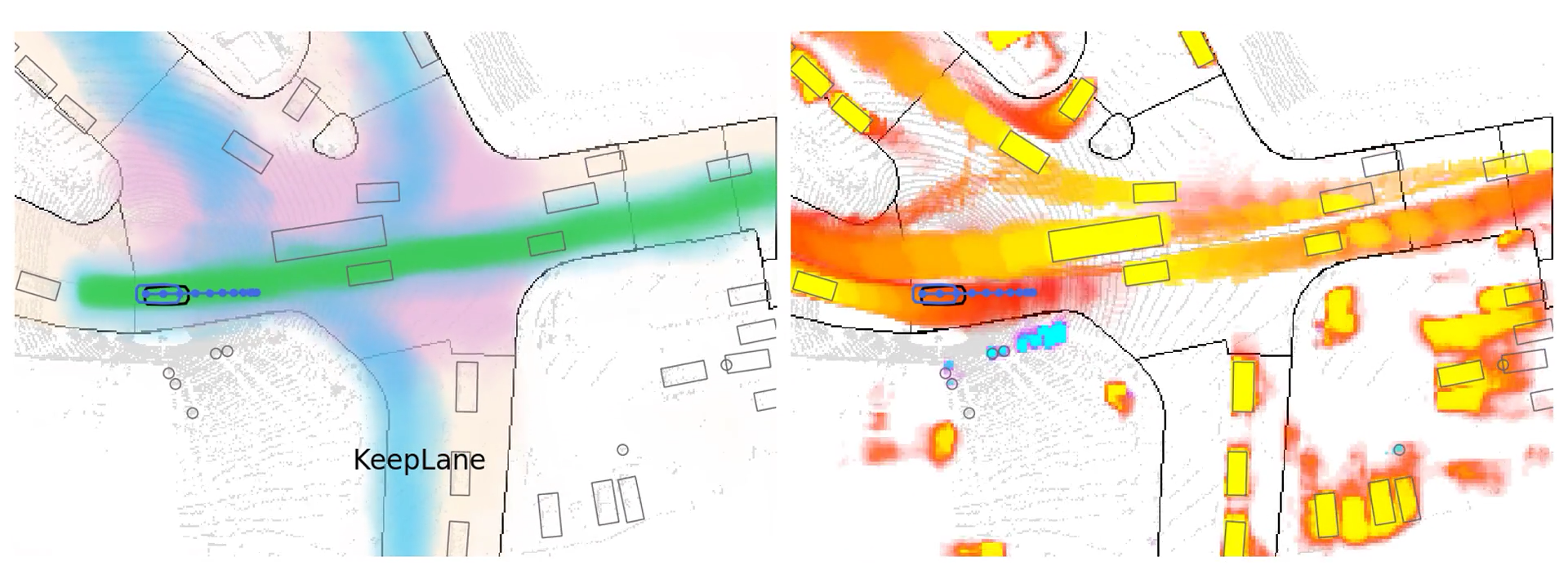} \\
        \includegraphics[width=0.8\linewidth]{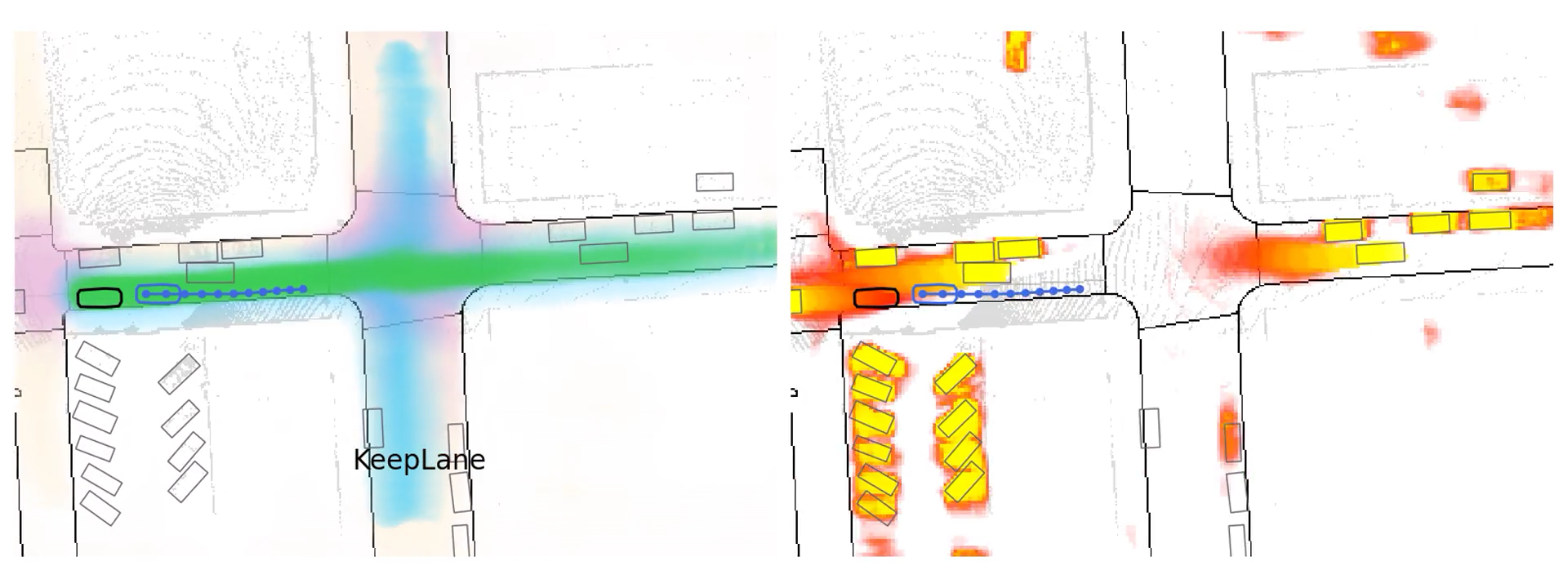} \\
        \includegraphics[width=0.8\linewidth]{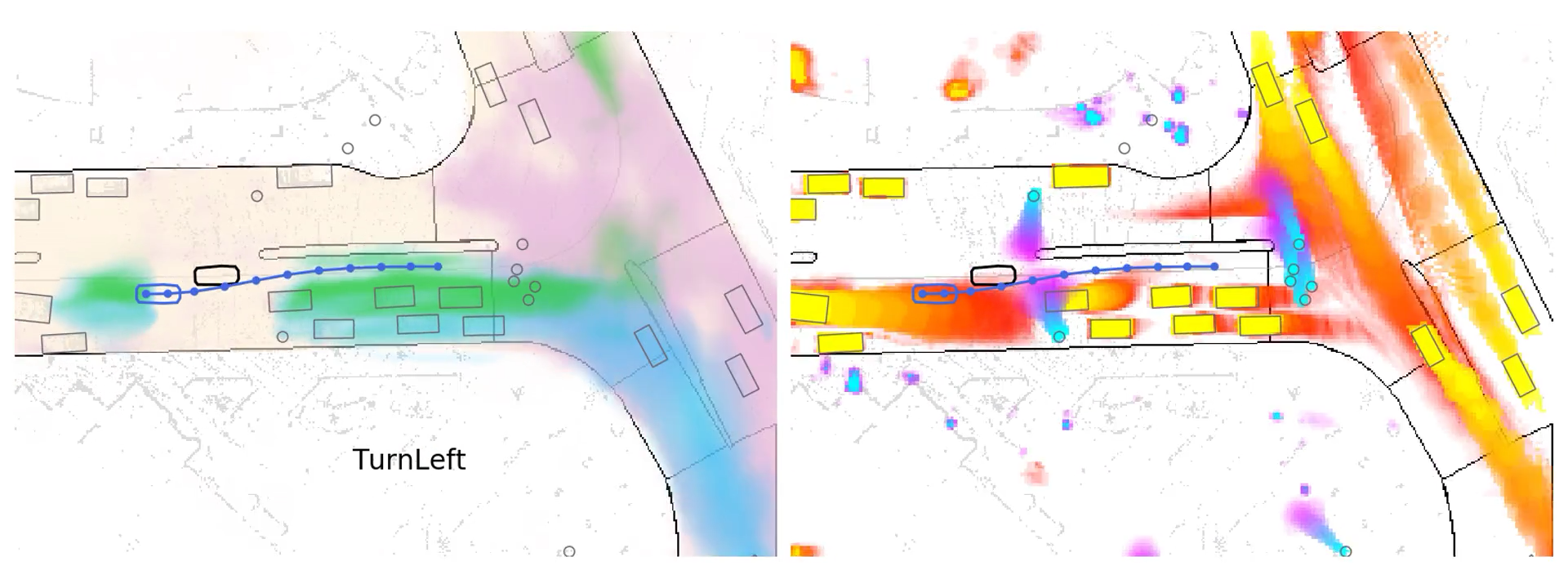} \\
        \includegraphics[width=0.8\linewidth]{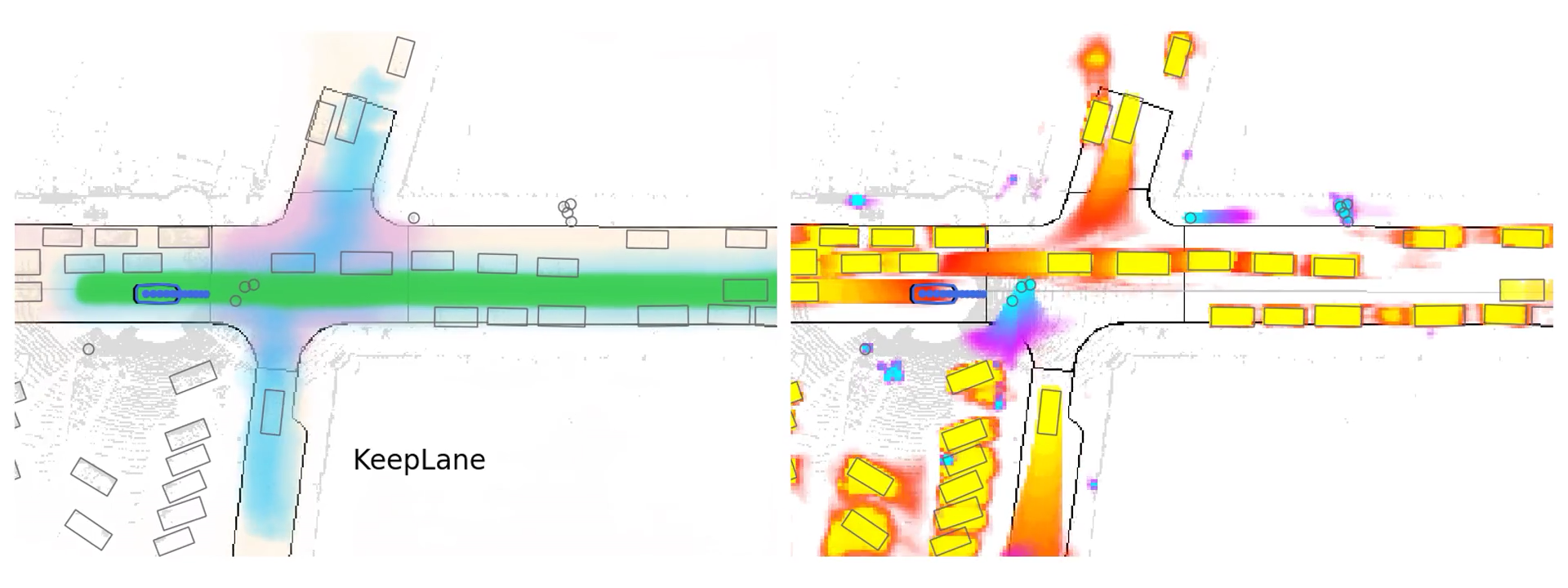} \\
    \end{tabular}
    \vspace{-10pt}
    \caption{\textbf{More qualitative results of MP3 in closed-loop.}}
    \label{fig:qualitative_supp2}
\end{figure*}

Figures~\ref{fig:qualitative_supp1}, \ref{fig:qualitative_supp2} showcase a solid understanding of both the static and dynamic  parts of the environment through the online map and dynamic occupancy predictions. These translate into good routings and safe maneuvers from the motion planner that are close to the expert demonstrations even after unrolling our own plans for several seconds and therefore deviating from the expert state.

\begin{figure*}[t]
    \vspace{-10pt}
    \centering
    \begin{tabular} {c}
        \includegraphics[width=0.8\linewidth]{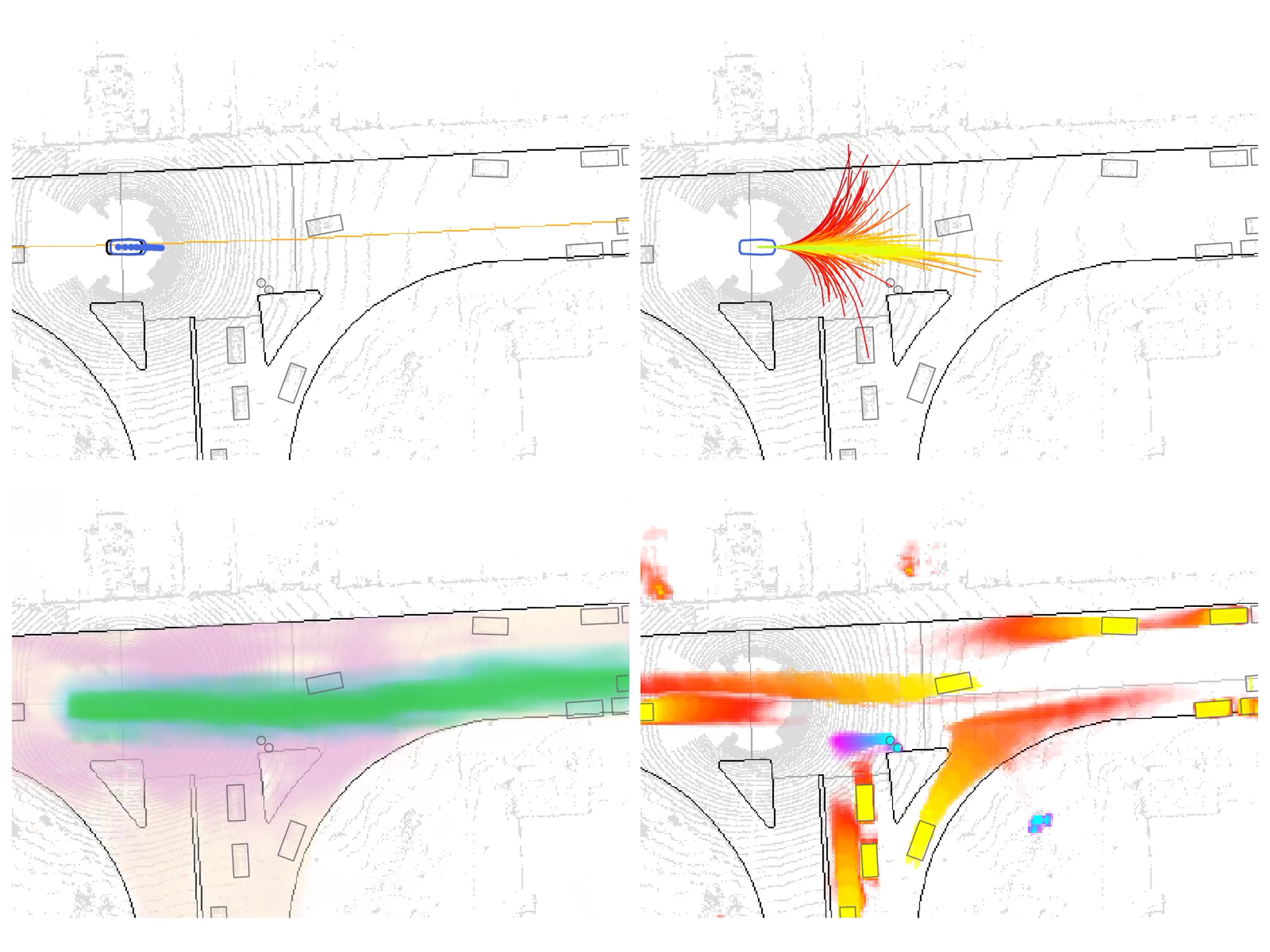} \\
    \end{tabular}
    \vspace{-10pt}
    \caption{\textbf{Yielding scenario at an intersection}. Top right showcases a visualization of the cost of the retrieved trajectory samples. The warmer the color the higher the cost.}
    \label{fig:qualitative_supp3}
\end{figure*}

\begin{figure*}[t]
    \vspace{-10pt}
    \centering
    \begin{tabular} {c}
        \includegraphics[width=0.8\linewidth]{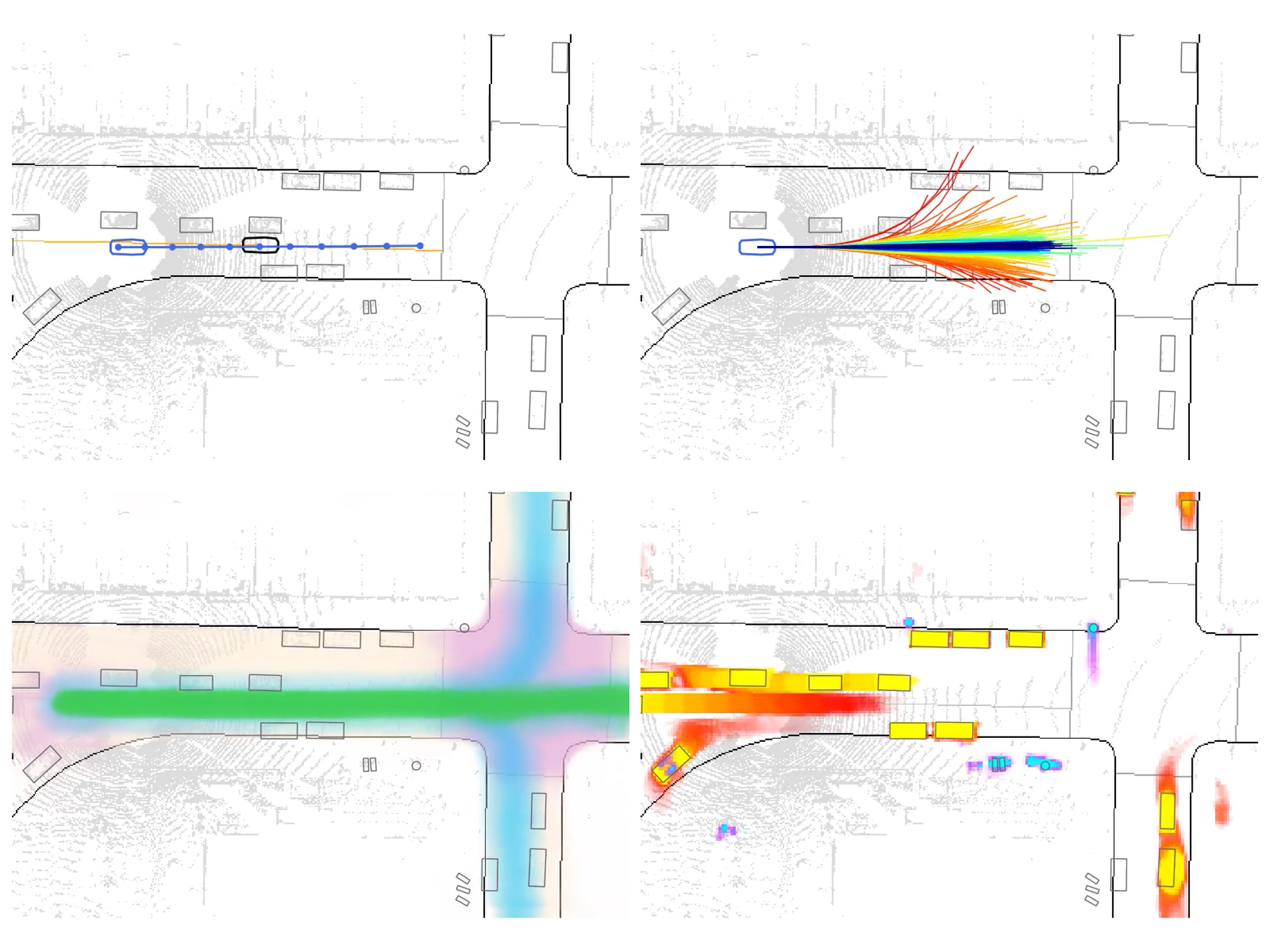} \\
    \end{tabular}
    \vspace{-10pt}
    \caption{\textbf{Cruising scenario}. As can be seen in the top right, the fast moving straight trajectories achieve the lowest score as there is no one in front of the SDV.}
    \label{fig:qualitative_supp4}
\end{figure*}

\begin{figure*}[t]
    \vspace{-10pt}
    \centering
    \begin{tabular} {c}
        \includegraphics[width=0.8\linewidth]{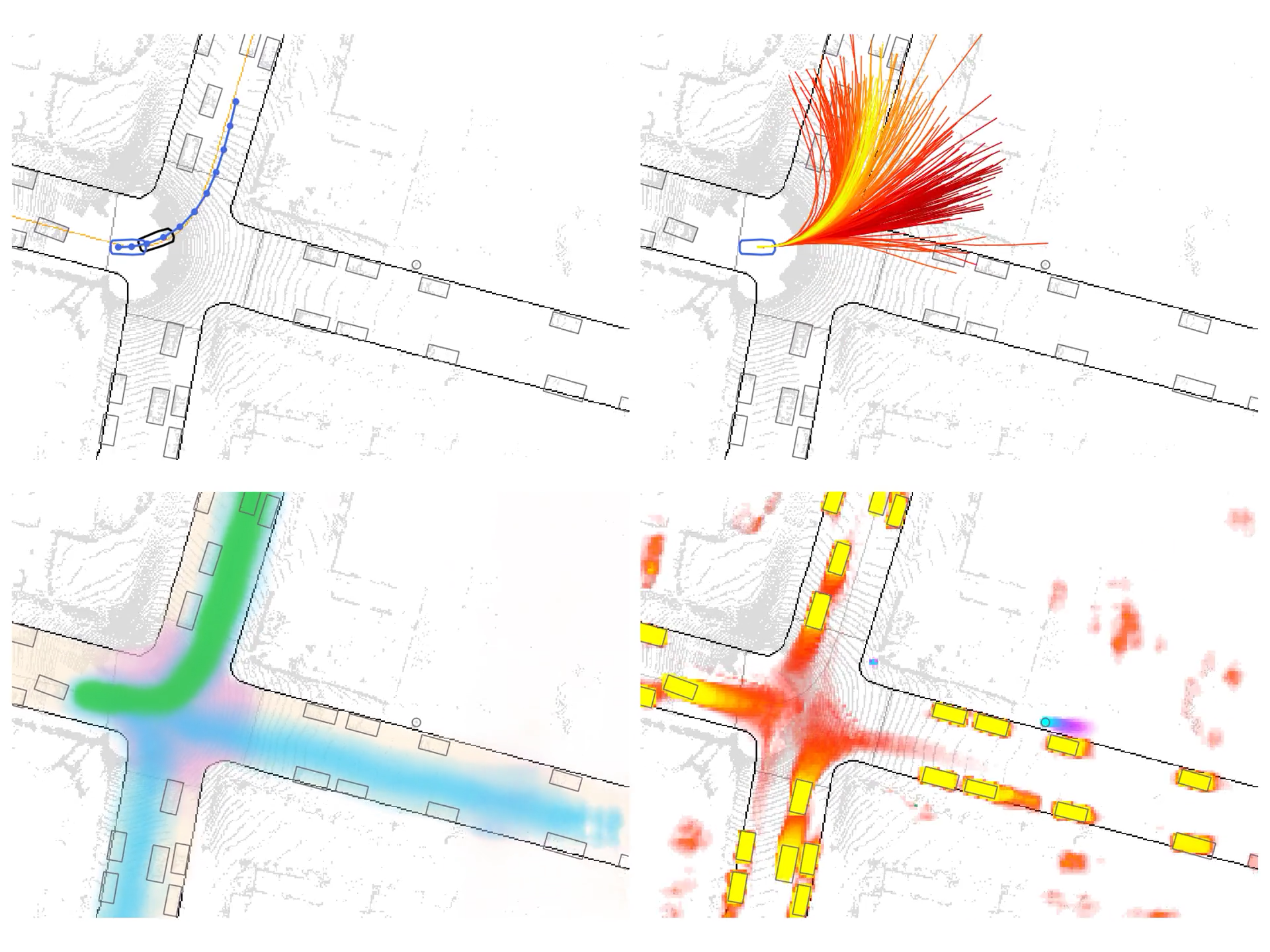} \\
    \end{tabular}
    \vspace{-10pt}
    \caption{\textbf{Left turn scenario} in which the SDV progresses fast.}
    \label{fig:qualitative_supp5}
\end{figure*}

\begin{figure*}[t]
    \vspace{-10pt}
    \centering
    \begin{tabular} {c}
        \includegraphics[width=0.8\linewidth]{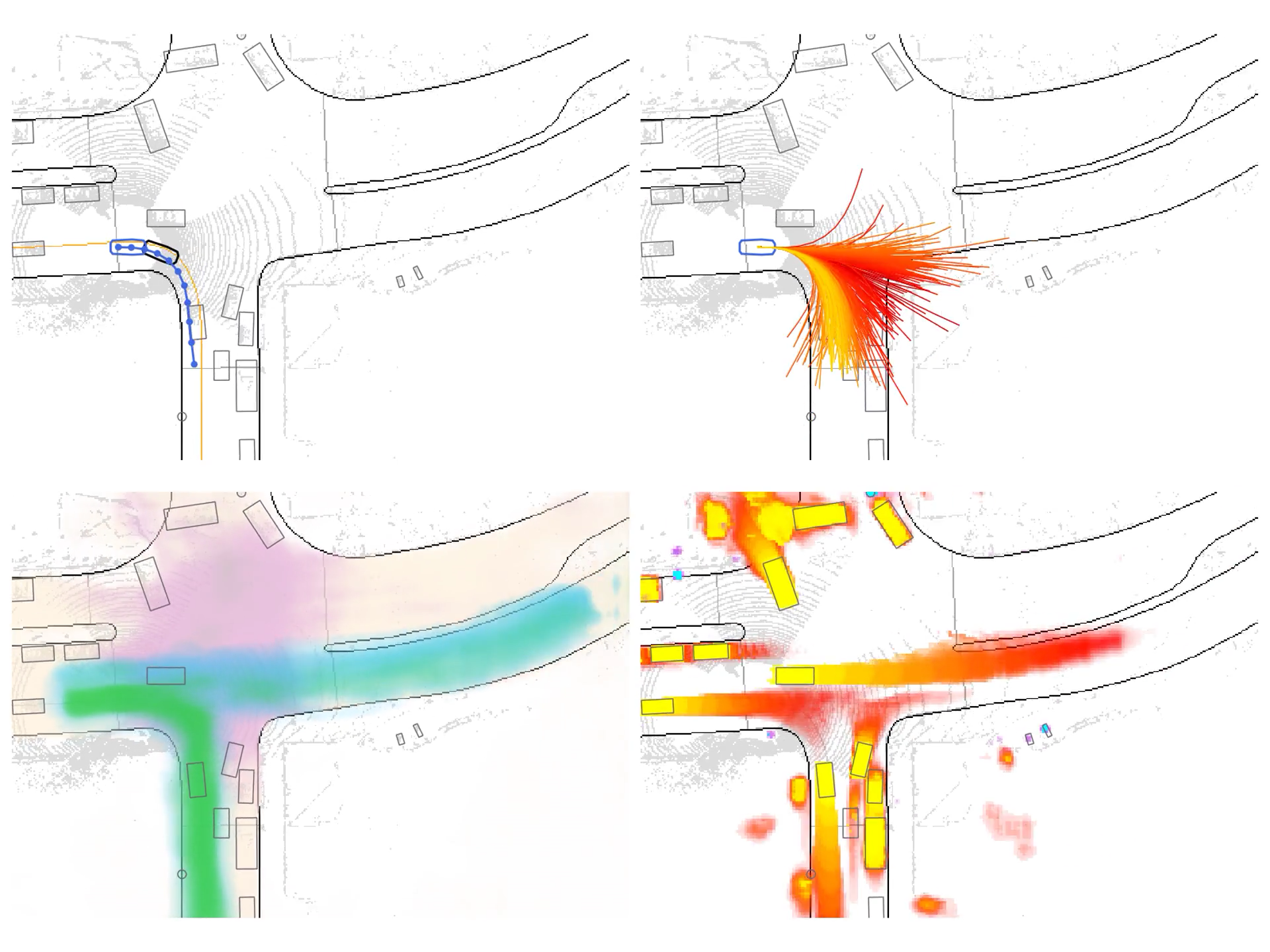} \\
    \end{tabular}
    \vspace{-10pt}
    \caption{\textbf{Right turn scenario.} We can see how the trajectory samples adapt to the current SDV velocity.}
    \label{fig:qualitative_supp6}
\end{figure*}
Moreover, Figs.~\ref{fig:qualitative_supp3}, \ref{fig:qualitative_supp4}, \ref{fig:qualitative_supp5}, \ref{fig:qualitative_supp6} provide further insight into the motion planner by additionally showing (a random subset of) the retrieved trajectory samples as well as their cost in a color map ranging from blue (lowest cost) to red (highest cost) in the top right image. The optimal trajectory (i.e. the one with the lowest cost) is plotted separately in the top left image.

\subsection{Plan comparison against baselines in closed-loop simulation}
\begin{figure*}[t]
    \vspace{-10pt}
    \centering
    \includegraphics[width=\textwidth, trim={0.0cm, 2.0cm, 2.0cm, 0.0cm}, clip]{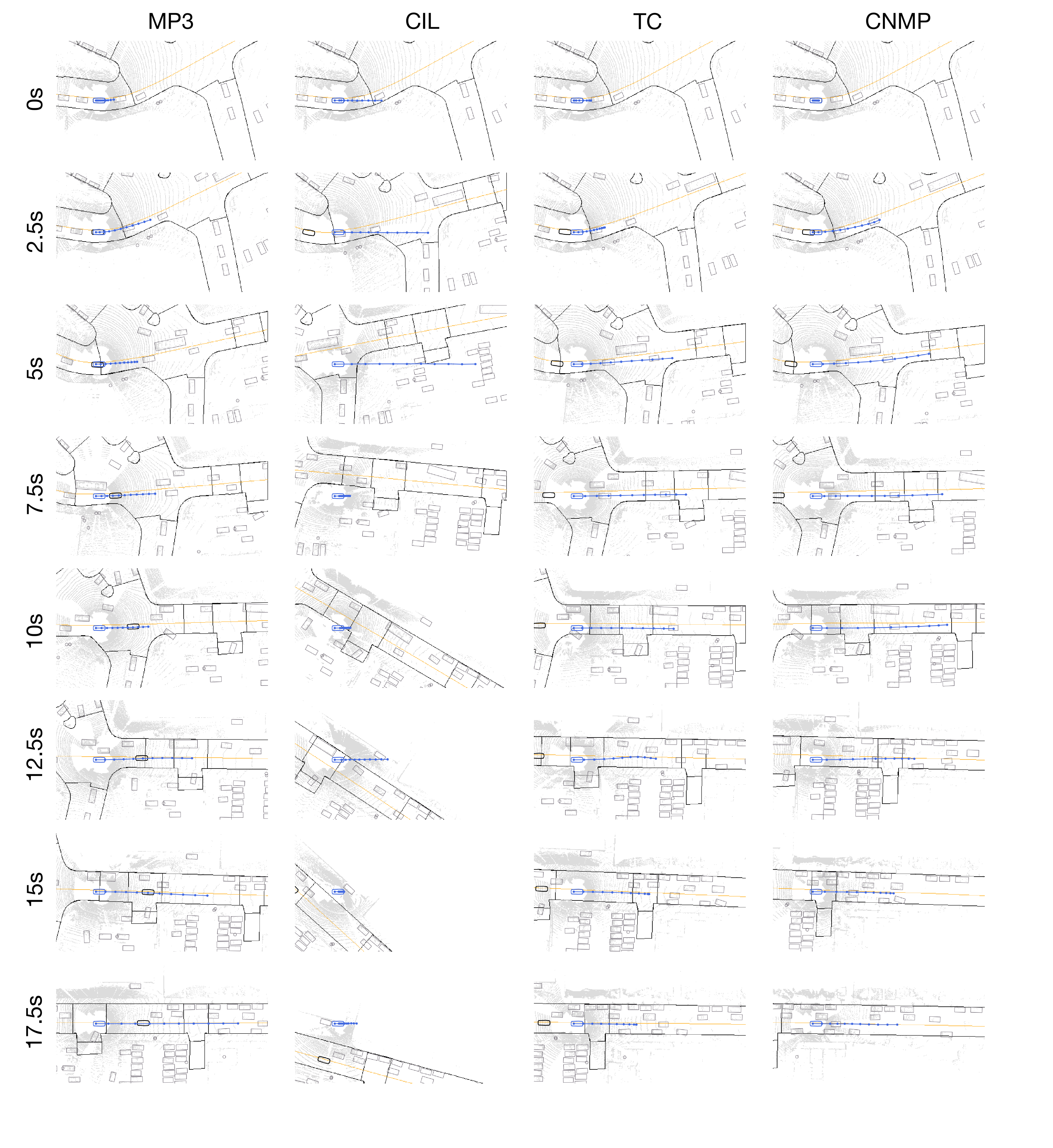}
    \caption{
        \textbf{Planning comparison in closed-loop} MP3 is the only method that follows the route and does not collide in this example.
    }
    \cutcaptiondown
    \label{fig:qualitative_planning0}
\end{figure*}

\begin{figure*}[t]
    \vspace{-10pt}
    \centering
    \includegraphics[width=\textwidth, trim={0.0cm, 2.0cm, 2.0cm, 0.0cm}, clip]{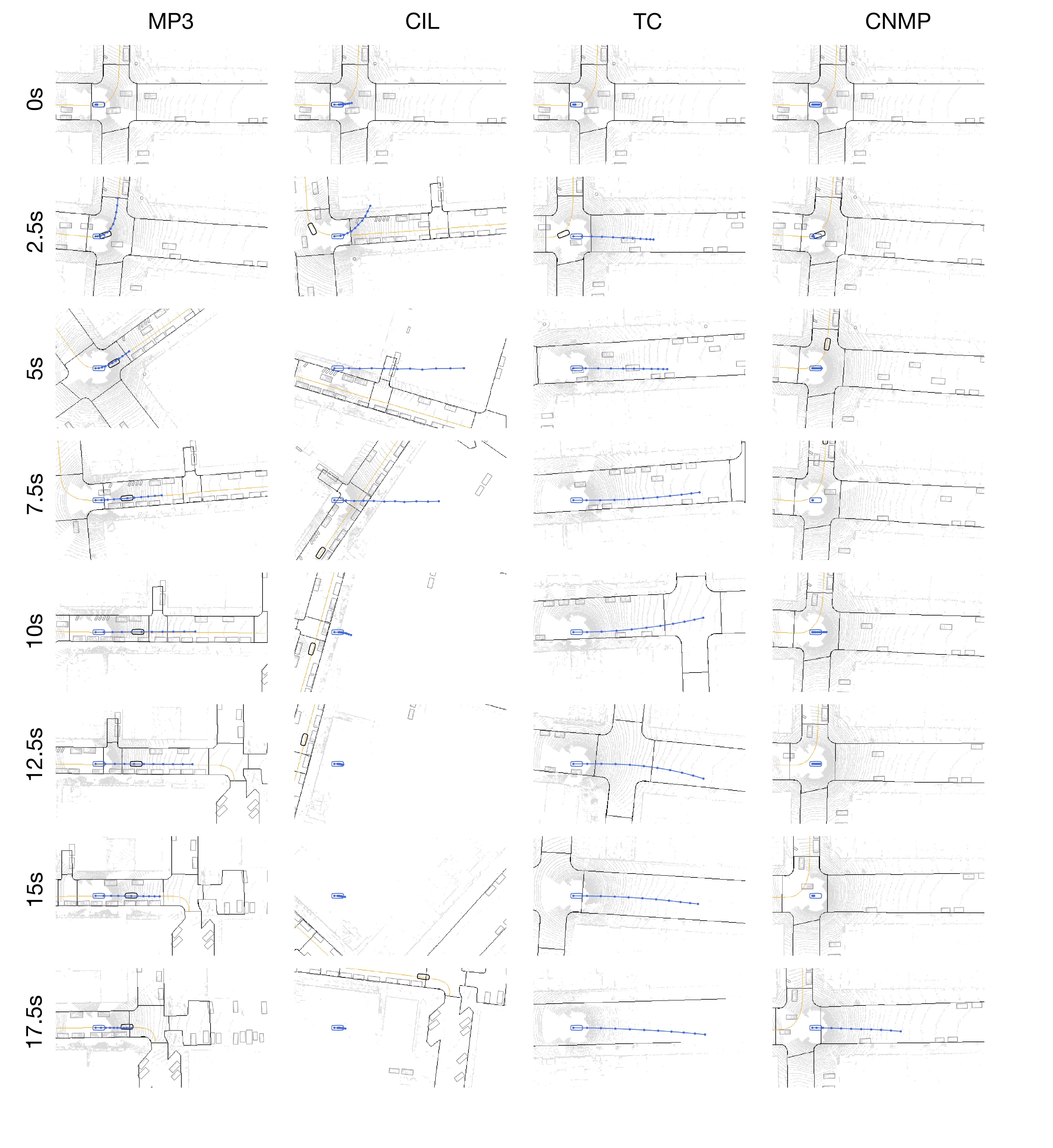}
    \caption{
        \textbf{More planning comparison in closed-loop}. 
        MP3 is the only method that manages to effectuate the unprotected left turn
    }
    \cutcaptiondown
    \label{fig:qualitative_planning1}
\end{figure*}

\begin{figure*}[t]
    \vspace{-10pt}
    \centering
    \includegraphics[width=\textwidth, trim={0.0cm, 2.0cm, 2.0cm, 0.0cm}, clip]{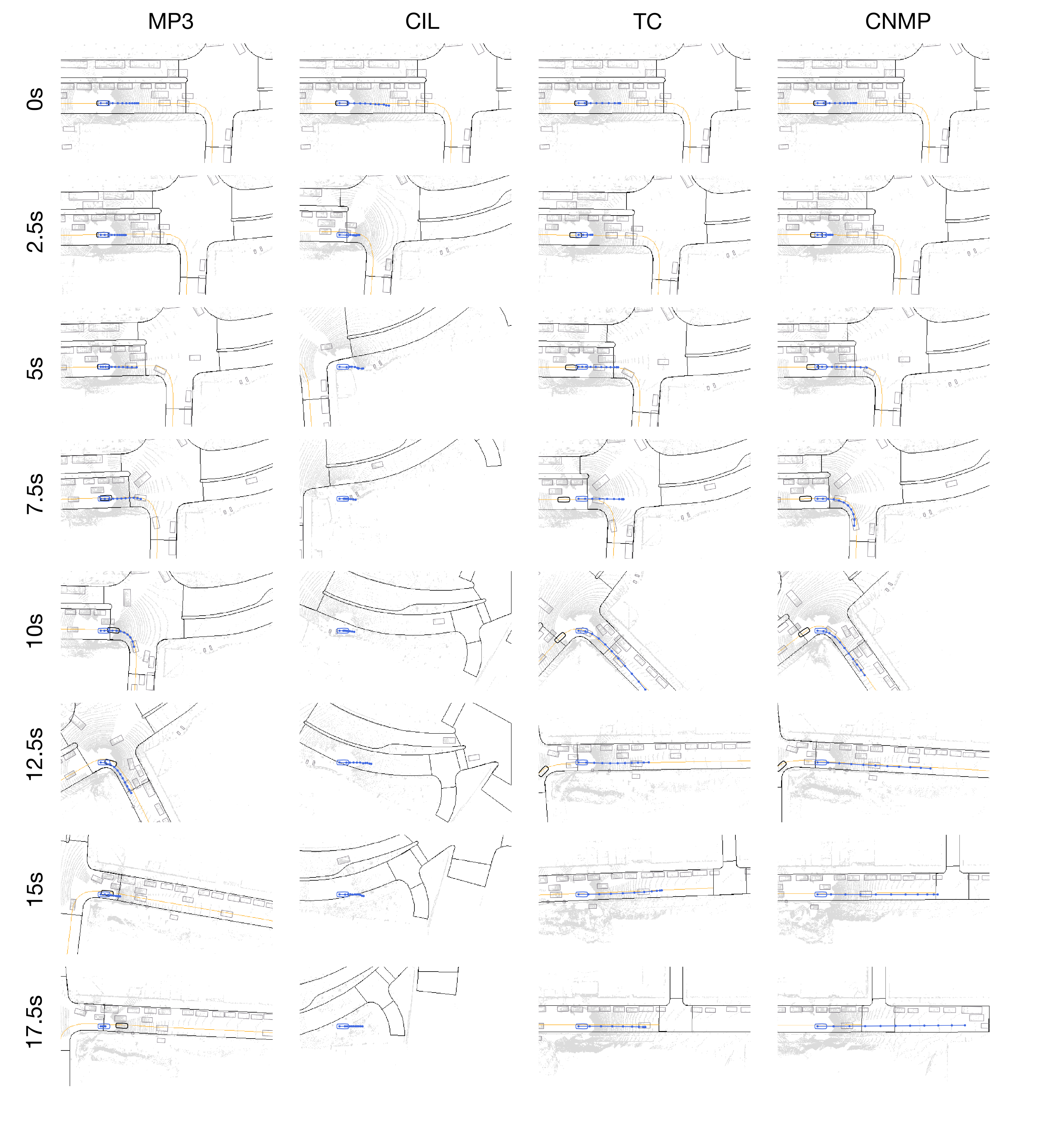}
    \caption{
        \textbf{More planning comparison in closed-loop}. 
        MP3 closely imitates the expert. CIL diverges from the route.  TC and CNMP stay on the route but collide with other vehicles.
    }
    \cutcaptiondown
    \label{fig:qualitative_planning2}
\end{figure*}

Figures~\ref{fig:qualitative_planning0}, \ref{fig:qualitative_planning1}, \ref{fig:qualitative_planning2} compare the plans from our method and those of the baselines in 3 different scenarios from our closed-loop simulations. 
In these figures, the expert driver state is shown in black for reference, and the plans in blue. 
The path that should be followed by obeying the high-level commands is shown in orange.
Each column is a method, and the rows represent a sequence of frames from a video (1 snapshot every 2.5 seconds of execution). 

\ifthenelse{\boolean{arxiv}}{
}{
    Please see our supplementary video for more comparisons.
} 

\end{document}